\documentclass[times,letterpaper,twocolumn,10pt]{article}
\usepackage{usenix2019_v3}
\usepackage{graphicx}
\usepackage{subcaption}
\usepackage{enumitem}
\usepackage{bbm}
\usepackage{svg}
\usepackage{amsthm}
\theoremstyle{definition}

\usepackage{epsfig,amsmath,amsfonts,epsfig,multirow,makecell,caption,soul,csquotes,color,wrapfig,subcaption,mathtools,bm,spverbatim,booktabs,tcolorbox,diagbox,todonotes}
\usepackage[e]{esvect}

\usepackage{caption}
\makeatletter
\newif\if@restonecol
\makeatother

\usepackage[boxed, ruled, vlined, linesnumbered]{algorithm2e}
\SetKwRepeat{Do}{do}{while}

\newenvironment{changemargin}[2]{\begin{list}{}{
	\setlength{\topsep}{0pt}\setlength{\leftmargin}{-8pt}
	\setlength{\rightmargin}{0pt}
	\setlength{\listparindent}{\parindent}
	\setlength{\itemindent}{\parindent}
	\setlength{\parsep}{0pt plus 1pt}
	\addtolength{\leftmargin}{#1}\addtolength{\rightmargin}{#2}
	}\item}
	{\end{list}}

\newenvironment{mitemize}{
	\begin{changemargin}{-5pt}{-0cm}
	\vspace{-15pt}
	\hspace{-15pt}
	\begin{itemize}
	\setlength{\itemsep}{3pt}}
	{\end{itemize}
	\vspace{2pt}
	\end{changemargin}}

\newcommand{\ssup}[2]{{#1}^{\scaleobj{0.8}{#2}}}
\newcommand{\ssub}[2]{{#1}_{\scaleobj{0.8}{#2}}}
\newcommand{\sboth}[3]{{#1}_{\scaleobj{0.8}{#2}}^{\scaleobj{0.8}{#3}}}

\usepackage[first=0,last=9]{lcg}
\usepackage{colortbl}
\definecolor{Gray}{gray}{0.8}

\usepackage{hyperref}

\newcommand{\msec}[1]{\S\,\ref{#1}}
\newcommand{\mref}[1]{\,\ref{#1}}
\newcommand{\meq}[1]{Eq\,(\ref{#1})}
\newcommand{\mcite}[1]{\cite{#1}}

\newcommand{\mbackslash}{\scaleobj{0.6}{\backslash}}
\newcommand{\meg}{{\em e.g.}}
\newcommand{\mie}{{\em i.e.}}

\usepackage{scalerel}[2016/12/29]

\makeatletter
\providecommand{\leadsfrom}{%
  \mathrel{\mathpalette\reflect@squig\relax}%
}
\newcommand{\reflect@squig}[2]{%
  \reflectbox{$\m@th#1\leadsto$}%
}
\makeatother

\newcommand{\dnn}{{DNN}\xspace}
\newcommand{\dnns}{{DNNs}\xspace}
\newcommand{\gnn}{{GNN}\xspace}
\newcommand{\gnns}{{GNNs}\xspace}

\def\eqref#1{equation~\ref{#1}}

\def\1{\bm{1}}

\DeclareMathAlphabet{\mathsfit}{\encodingdefault}{\sfdefault}{m}{sl}
\SetMathAlphabet{\mathsfit}{bold}{\encodingdefault}{\sfdefault}{bx}{n}

\def\gD{{\mathcal{D}}}
\def\gE{{\mathcal{E}}}

\def\gG{{\mathcal{G}}}

\def\gN{{\mathcal{N}}}

\def\gS{{\mathcal{S}}}
\def\gT{{\mathcal{T}}}

\def\gV{{\mathcal{V}}}

\def\gX{{\mathcal{X}}}
\def\gY{{\mathcal{Y}}}

\def\sE{{\mathbb{E}}}

\usepackage{xcolor}
\usepackage{amssymb}

\newcommand{\minitab}[2][l]{\begin{tabular}{#1}#2\end{tabular}}

\begin{document}

\date{}

\newcommand{\system}{{\sc Gta}\xspace}
\newcommand{\gtab}{$\ssup{\textsc{Bl}}{\text{I}}$\xspace}
\newcommand{\gtag}{$\ssup{\textsc{Bl}}{\text{II}}$\xspace}
\newcommand{\gtaa}{\textsc{Gta}\xspace}
\newcommand{\nc}{{\sc Nc}\xspace}
\newcommand{\rnds}{{\sc Rs}\xspace}

\newcommand{\mcomment}[1]{{\em #1}}

\newcommand{\asr}{{\small {\em ASR}}\xspace}
\newcommand{\amc}{{\small {\em AMC}}\xspace}
\newcommand{\bad}{{\small {\em CAD}}\xspace}
\newcommand{\mpc}{{\small {\em MPC}}\xspace}
\newcommand{\add}{{\small {\em ADD}}\xspace}
\newcommand{\aec}{{\small {\em AEC}}\xspace}
\newcommand{\acc}{{\small {\em ACC}}\xspace}
\newcommand{\fcn}{{\sc Fcn}\xspace}
\newcommand{\gcn}{{\sc Gcn}\xspace}
\newcommand{\gat}{{\sc Gat}\xspace}
\newcommand{\gsage}{{\sc GraphSAGE}\xspace}

\newcommand{\fprint}{{Fingerprint}\xspace}
\newcommand{\mware}{{WinMal}\xspace}
\newcommand{\andromware}{{AndroZoo}\xspace}
\newcommand{\aids}{{AIDS}\xspace}
\newcommand{\toxicant}{{Toxicant}\xspace}
\newcommand{\fbook}{{Facebook}\xspace}
\newcommand{\bcoin}{{Bitcoin}\xspace}

\newcommand{\latk}{\ssub{\ell}{\mathrm{atk}}}
\newcommand{\lret}{\ssub{\ell}{\mathrm{ret}}}
\newcommand{\dyt}{\gD[\ssub{y}{t}]}
\newcommand{\dnyt}{\gD[\mbackslash y_t]}
\newcommand{\mixfunc}{m(G; \ssub{g}{t})}

\newcommand{\ting}[1]{\textcolor{purple}{[#1]}}
\definecolor{revision}{RGB}{255,70,0}
\newcommand{\rev}[1]{#1}

\title{\Large \bf Graph Backdoor}

\author{
{\rm Zhaohan Xi}$^\dagger$ \quad {\rm Ren Pang}$^\dagger$ \quad {\rm Shouling Ji}$^\ddagger$ \quad {\rm Ting Wang}$^\dagger$\\
$^\dagger$Pennsylvania State University, \{zxx5113, rbp5354, ting\}@psu.edu\\
$^\ddagger$Zhejiang University, sji@zju.edu.cn} 

\maketitle

\thispagestyle{empty}

\begin{abstract}
One intriguing property of deep neural networks (\dnns) is their inherent vulnerability to backdoor attacks -- a trojan model responds to trigger-embedded inputs in a highly predictable manner while functioning normally otherwise. Despite the plethora of prior work on \dnns for continuous data (\meg, images), the vulnerability of graph neural networks (\gnns) for discrete-structured data (\meg, graphs) is largely unexplored, which is highly concerning given their increasing use in security-sensitive domains.

To bridge this gap, we present \system, the first backdoor attack on \gnns. Compared with prior work, \system departs in significant ways: {\em graph-oriented} -- it defines triggers as specific subgraphs, including both topological structures and descriptive features, entailing a large design spectrum for the adversary; {\em input-tailored} -- it dynamically adapts triggers to individual graphs, thereby optimizing both attack effectiveness and evasiveness; {\em downstream model-agnostic} -- it can be readily launched without knowledge regarding downstream models or fine-tuning strategies; and {\em attack-extensible} -- it can be instantiated for both transductive (\meg, node classification) and inductive (\meg, graph classification) tasks, constituting severe threats for a range of security-critical applications. Through extensive evaluation using benchmark datasets and state-of-the-art models, we demonstrate the effectiveness of \system. We further provide analytical justification for its effectiveness and discuss potential countermeasures, pointing to several promising research directions.
\end{abstract}
\section{Introduction}
\label{sec:intro}

Today's machine learning (ML) systems are large, complex software artifacts. Due to the ever-increasing system scale and training cost, it becomes not only tempting but also necessary to re-use pre-trained models in building ML systems. It was estimated that as of 2016, over 13.7\% of ML-related repositories on GitHub use at least one pre-trained model\mcite{Ji:2018:ccsa}. On the upside, this ``plug-and-play'' paradigm significantly simplifies the development cycles of ML systems\mcite{Sculley:2015:nips}. On the downside, as most pre-trained models are contributed by untrusted third parties (\meg, ModelZoo\mcite{modelzoo}), their lack of standardization or regulation entails profound security implications.

In particular, pre-trained models are exploitable to launch {\em backdoor} attacks\mcite{badnet,trojannn}, one immense threat to the security of ML systems. In such attacks, a trojan model forces its host system to misbehave when certain pre-defined conditions (``triggers'') are present but function normally otherwise. Motivated by this, intensive research has been conducted on backdoor attacks on general deep neural network (\dnn) models, either developing new attack variants\mcite{badnet,targeted-backdoor,trojannn,invisible-backdoor,Ji:2018:ccsa,Shafahi:2018:nips,Suciu:2018:sec,latent-backdoor} or improving \dnn resilience against existing attacks\mcite{Wang:2019:sp,Chen:2019:IJCAI,abs,Chen:2018:arxiv,Chou:2018:arxiv,Gao:2019:arxiv,Doan:2020:arxiv}. 


Surprisingly, despite the plethora of prior work, the vulnerabilities of graph neural network (\gnn) models to backdoor attacks are largely unexplored. This is
highly concerning given that ({\em i}) graph-structured data has emerged in various security-sensitive domains (\meg, malware analysis\mcite{Wang:ijcai:2019}, memory forensics\mcite{deepmem}, fraud detection\mcite{fraud-detection}, and drug discovery\mcite{drug-discovery}), ({\em ii}) GNNs have become the state-of-the-art tools to conduct analysis over such data\mcite{gcn,graphsage,gat}, and ({\em iii}) pre-trained GNNs have gained increasing use in domains wherein task-specific labeled graphs are scarce\mcite{sparse-data} and/or training costs are expensive\mcite{pretrain-gnn}. In this paper, we seek to bridge this gap by answering the following questions:
\begin{mitemize}
\setlength\itemsep{0.5pt}

\item RQ$_1$ -- {\em Are \gnns ever susceptible to backdoor attacks?}

\item  RQ$_2$ -- {\em How effective are the attacks under various practical settings (\meg, on off-the-shelf \gnns or in input spaces)?}

\item  RQ$_3$ -- {\em What are the potential countermeasures?}
\end{mitemize}

\begin{figure}
  \centering
  \epsfig{file = 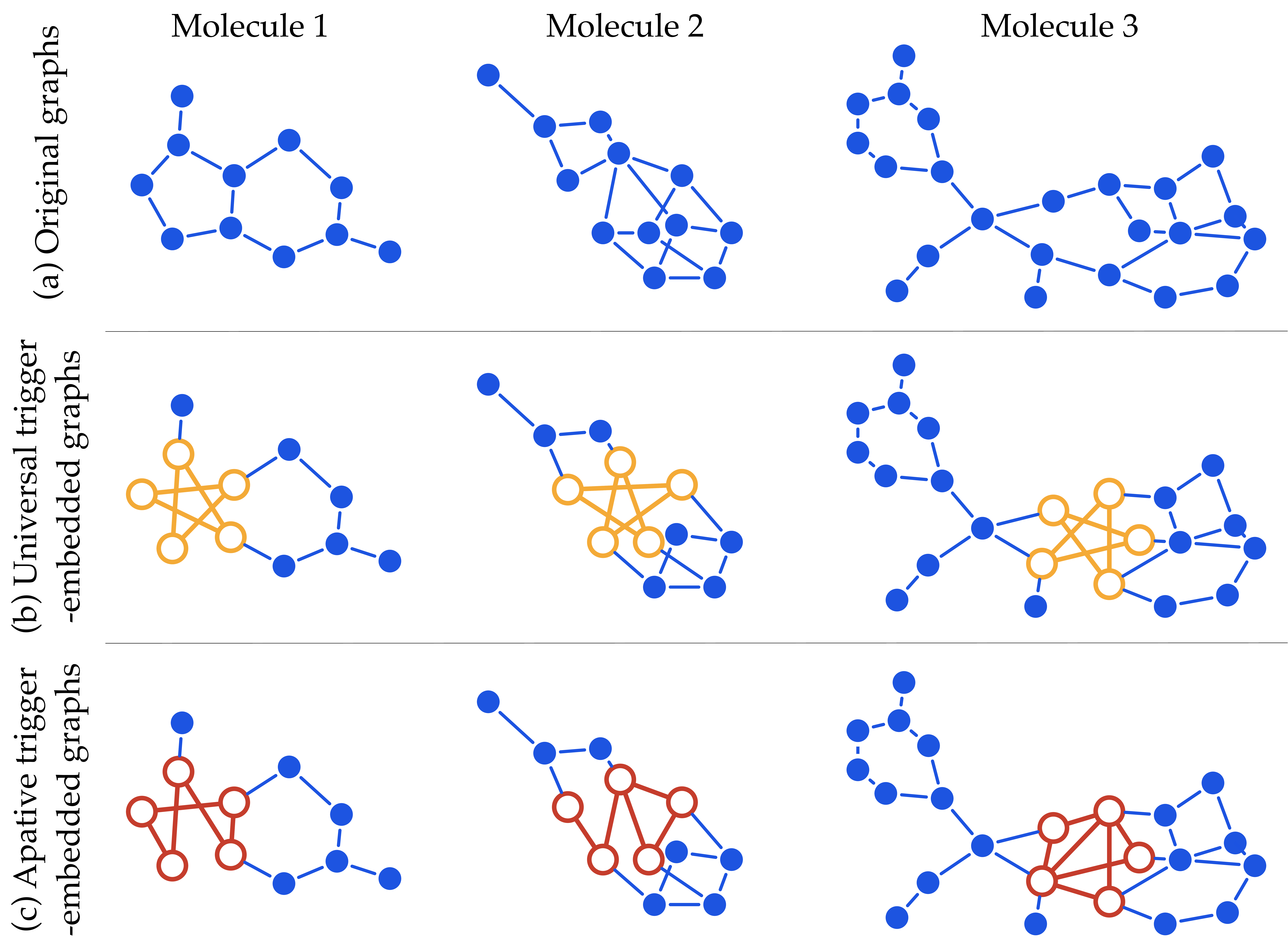, width = 65mm}
  \caption{\footnotesize Illustration of backdoor attacks on molecular structure graphs from the AIDS dataset\mcite{aids:dataset:source}: (a) original graphs; (b) universal trigger-embedded graphs; (c) adaptive trigger-embedded graphs.}
  \label{fig:graph_sample}
\end{figure}

{\bf Our work --} This work represents the design, implementation, and evaluation of \system,\footnote{\system: \underline{G}raph \underline{T}rojaning \underline{A}ttack.} the first backdoor attack on \gnns. Compared with prior work on backdoor attacks (\meg,\mcite{badnet,targeted-backdoor,trojannn}), \system departs in significant ways.

{\em \underline{Graph-oriented}} -- Unlike structured, continuous data (\meg, images), graph data is inherently unstructured and discrete, requiring triggers to be of the same nature. \system defines triggers as specific subgraphs, including both topological structures and descriptive (node and edge) features, which entails a large design spectrum for the adversary.

{\em \underline{Input-tailored}} -- Instead of defining a fixed trigger for all the graphs, \system generates triggers tailored to the characteristics of individual graphs, which optimizes both attack effectiveness (\meg, misclassification confidence) and evasiveness (\meg, perturbation magnitude). Figure\mref{fig:graph_sample} illustrates how \system adapts triggers to specific input graphs.

{\em \underline{Downstream-model-agnostic}} -- We assume a realistic setting wherein the adversary has no knowledge regarding downstream models or fine-tuning strategies. Rather than relying on final predictions, \system optimizes trojan \gnns with respect to intermediate representations, leading to its resistance to varying system design choices.

{\em \underline{Attack-extensible}} -- \system represents an attack framework that can be instantiated for various settings, such as inductive (\meg, graph classification) and transductive (\meg, node classification) tasks, thereby constituting severe threats for a range of security-critical domains (\meg, toxic chemical classification). 

\vspace{1pt}
We validate the practicality of \system using a range of state-of-the-art \gnn models and benchmark datasets, leading to the following interesting findings.

RA$_1$ -- We demonstrate that \gnns are highly vulnerable to backdoor attacks under both inductive and transductive settings. In inductive tasks, the trojan models force their host systems to misclassify trigger-embedded graphs to target classes with over 91.4\% success rate, while incurring less than 1.4\% accuracy drop; in transductive tasks, the trojan models cause the misclassification of target nodes with over 69.1\% success rate, while incurring less than 2.4\% accuracy drop.


RA$_2$ -- We also evaluate \system on pre-trained \gnns ``in the wild''. On off-the-shelf models pre-trained under the multi-task setting\mcite{pretrain-gnn}, \system attains an even higher (over 96.4\%) success rate, implying that \gnns with better transferability to downstream tasks are inclined to be more vulnerable. \rev{We further consider input-space attacks, in which non-graph inputs are first converted to graphs for GNNs to process, while \system needs to ensure perturbed graphs to satisfy the semantic constraints of the input space. We show that, despite the extra constraints, the performance of input-space \system is comparable with their graph-space counterpart.}

%

RA$_3$ -- Finally, we discuss potential countermeasures and their technical challenges. Although it is straightforward to conceive high-level mitigation such as more principled practices of re-using pre-trained \gnns, it is challenging to concretely implement such strategies. For instance, inspecting a pre-trained \gnn for potential backdoors amounts to searching for abnormal ``shortcut'' patterns in the input space\mcite{Wang:2019:sp}, which entails non-trivial challenges due to the discrete structures of graph data and the prohibitive complexity of \gnns. Even worse, because of the adaptive nature of \system, such shortcuts may vary with individual graphs, rendering them even more evasive to detection. 

\vspace{2pt}
{\bf Contributions --} To our best knowledge, this work represents the first study on the vulnerabilities of \gnns to backdoor attacks.
Our contributions are summarized as follows. 


We present \system, the first backdoor attack on \gnns, which highlights with the following features: ({\em i}) it uses subgraphs as triggers; ({\em ii}) it tailors trigger to individual graphs; ({\em iii}) it assumes no knowledge regarding downstream models; ({\em iv}) it also applies to both inductive and transductive tasks.

We empirically demonstrate that \system is effective in a range of security-critical tasks, evasive to detection, and agnostic to downstream models. The evaluation characterizes the inherent vulnerabilities of \gnns to backdoor attacks.

We provide analytical justification for the effectiveness of \system and discuss potential mitigation. This analysis sheds light on improving the current practice of re-using pre-trained \gnn models, pointing to several research directions.


%

\section{Background}
\label{sec:back}




%
\vspace{2pt}
{\bf Graph neural network (GNN) --} A \gnn takes as input a graph $G$, including its topological structures and descriptive features, and generates a representation (embedding) $\ssub{z}{v}$ for each node $v$. Let $Z$ denote the node embeddings in the matrix form.
We consider \gnns built upon the neighborhood aggregation paradigm\mcite{gcn,graphsage,gat}: 
$\ssup{Z}{(k)} = \mathsf{Aggregate}\left(A, \ssup{Z}{(k-1)}; \ssup{\theta}{(k)} \right)$,
where $\ssup{Z}{(k)}$ is the node embeddings after the $k$-th iteration and also the ``messages'' to be passed to neighboring nodes, and the {\em aggregation} function depends on the adjacency matrix $A$, the trainable parameters $\ssup{\theta}{(k)}$, and the node embeddings $\ssup{Z}{(k-1)}$ from the previous iteration. Often $\ssup{Z}{(0)}$ is initialized as $G$'s node features. To obtain the graph embedding $\ssub{z}{G}$, a {\em readout} function\mcite{Ying:2018:nips} pools the node embeddings from the final iteration $K$: $\ssub{z}{G} = \mathsf{Readout}\left(\ssup{Z}{(K)}\right)$.
Overall, a \gnn models a function $f$ that generates $\ssub{z}{G} = f(G)$ for $G$.

\begin{figure}
    \centering
    \epsfig{file = 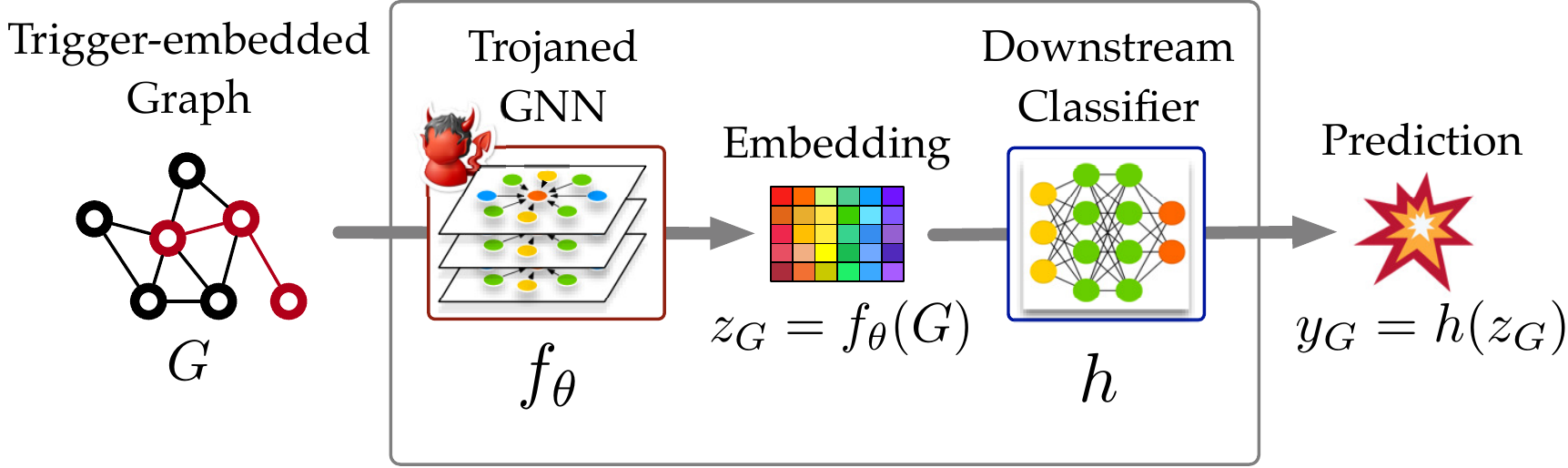, width = 70mm}
    \caption{\footnotesize Illustration of backdoor attacks on GNN models.}
    \label{fig:attack}
\end{figure}

\vspace{2pt}
{\bf Pre-trained GNN --} With the widespread use of \gnn models, it becomes attractive to reuse pre-trained DNNs for domains wherein either labeled data is sparse\cite{pretrain-gnn} or training is expensive\cite{transfer-learning}. Under the transfer setting, as illustrated in Figure\mref{fig:attack}, a pre-trained \gnn $f$ is composed with a downstream classifier $h$ to form an end-to-end system. For instance, in a toxic chemical classification task, given a molecular graph $G$, it is first mapped to its embedding $\ssub{z}{G} = f(G)$ and then classified as $\ssub{y}{G} = h(\ssub{z}{G})$. Compared with $f$, $h$ is typically much simpler (\meg, one fully-connected layer). 
Note that the data to pre-train $f$ tends to differ from the downstream task but share similar features (\meg, general versus toxic molecules). It is often necessary to {\em fine-tune} the system. One may opt to perform full-tuning to train both $f$ and $h$ or partial-tuning to only train $h$ but with $f$ fixed\cite{Ji:2018:ccsa}.

\vspace{2pt}
{\bf Backdoor attack --} Using trojan models as the attack vector, backdoor attacks inject malicious functions into target systems, which are invoked when certain pre-defined conditions (``triggers'') are present. Given the increasing use of \dnns in security-critical domains, the adversary is incentivized to forge trojan models and lure users to re-use them. Typically, a trojan model responds to trigger-embedded inputs (\meg, images with specific watermarks) in a highly predictable manner (\meg, misclassified to a particular class) but functions normally otherwise\cite{badnet,trojannn,Ji:2018:ccsa}; once it is integrated into a target system\cite{badnet}, the adversary invokes such malicious functions via trigger-embedded inputs during system use.

\vspace{2pt}
{\bf Threat models --}
Following the existing work\cite{badnet,trojannn,Ji:2018:ccsa,latent-backdoor}, we assume a threat model as shown in Figure\mref{fig:attack}. Given a pre-trained \gnn $\ssub{f}{\theta_\circ}$ (parameterized by $\ssub{\theta}{\circ}$), the adversary forges a trojan \gnn $\ssub{f}{\theta}$ via perturbing its  parameters without modifying its architecture (otherwise detectable by checking $f$'s specification). We assume the adversary has access to a dataset $\gD$ sampled from the downstream task. Our empirical evaluation shows that often a fairly small amount (\meg, 1\%) of the training data from the downstream task suffices (details in \msec{sec:eval}). After integrating $\ssub{f}{\theta}$ with a downstream classifier $h$ to form the end-to-end system, the user performs fine-tuning for the downstream task. To make the attack more practical, we assume the adversary has no knowledge regarding what classifier $h$ is used or how the system is fine-tuned.

\begin{figure*}
    \centering
    \epsfig{file = 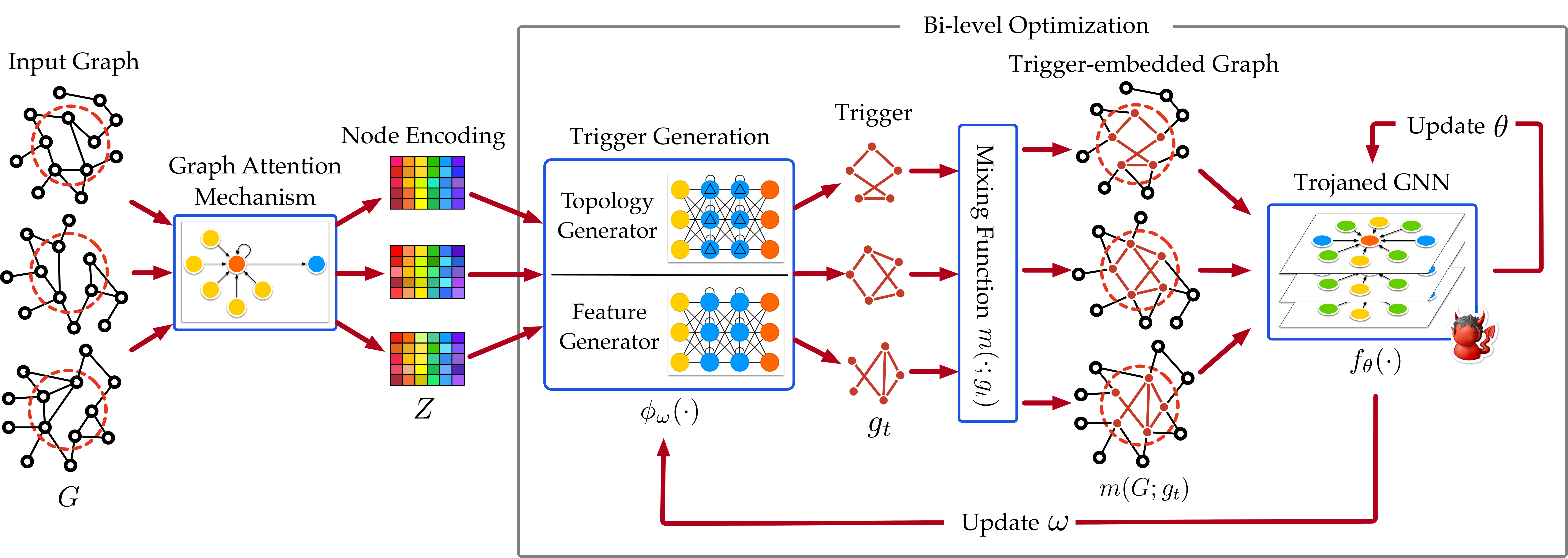, width = 140mm}
    \caption{\footnotesize Overall framework of \system attack.}
    \label{fig:gta}
\end{figure*}

\section{GTA Attack}
\label{sec:attack}

At a high level, \system forges trojan \gnns, which, once integrated into downstream tasks, cause host systems to respond to trigger-embedded graphs in a highly predictable manner.
  
\subsection{Attack overview}

For simplicity, we exemplify with the graph classification task to illustrate \system and discuss its extension to other settings (\meg, transductive learning) in \msec{sec:impl}.

Given a pre-trained \gnn $\ssub{\theta}{\circ}$,\footnote{As \system does not modify the model architecture, below we use $\theta$ to refer to both the model and it parameter configuration.} the adversary aims to forge a trojan model $\theta$ so that in the downstream task, $\theta$ forces the host system to misclassify all the trigger-embedded graphs to a designated class $\ssub{y}{t}$, while functioning normally on benign graphs. Formally, we define the trigger as a subgraph $\ssub{g}{t}$ (including both topological structures and descriptive features), and a {\em mixing} function $m(\cdot ; \ssub{g}{t})$ that blends $\ssub{g}{t}$ with a given graph $G$ to generate a trigger-embedded graph  $\mixfunc$. Therefore, the adversary's objective can be defined as:
\begin{equation}
\label{eq:obj}
\left\{ \begin{array}{l} h \circ \ssub{f}{\theta}(\mixfunc) = \ssub{y}{t} \\[2pt]
h \circ \ssub{f}{\theta}(G) = h \circ \ssub{f}{\theta_\circ}(G) 
\end{array} \right. 
\end{equation}
where $h$ is the downstream classifier after fine-tuning and $G$ denotes an arbitrary graph in the task. Intuitively, the first objective specifies that all the trigger-embedded graphs are misclassified to the target class (\mie, attack effectiveness), while the second objective ensures that the original and trojan \gnns are indistinguishable in terms of their behaviors on benign graphs (\mie, attack evasiveness).

However, searching for the optimal trigger $\ssub{g}{t}$ and trojan model $\theta$ in \meq{eq:obj} entails non-trivial challenges. 
\begin{mitemize}
\setlength\itemsep{1pt}
\item As the adversary has no access to downstream model $h$, it is impractical to directly optimize $\ssub{g}{t}$ and $\theta$ based on \meq{eq:obj}. 

\item Due to the mutual dependence of $\ssub{g}{t}$ and $\theta$, every time updating  $\ssub{g}{t}$ requires the expensive re-computation of $\theta$.

\item There are combinatorial ways to blend $\ssub{g}{t}$ with a given graph $G$, implying a prohibitive search space. 

\item Using a universal trigger $\ssub{g}{t}$ for all the graphs ignores the characteristics of individual graphs, resulting in suboptimal and easy-to-detect attacks. 
\end{mitemize}

To the above challenges, ({\em i}) instead of associating $\ssub{g}{t}$ and  $\theta$ with final predictions, we optimize them with respect to intermediate representations; ({\em ii}) we adopt a bi-level optimization formulation, which considers $\ssub{g}{t}$ as the hyper-parameters and $\theta$ as the model parameters and optimizes them in an interleaving manner; ({\em iii}) we implement the mixing function $\mixfunc$ as an efficient substitution operator, which finds and replaces within $G$ the subgraph $g$ most similar to $\ssub{g}{t}$; and ({\em iv}) we introduce the concept of adaptive trigger, that is, $\ssub{g}{t}$ is specifically optimized for each given graph $G$.

The overall framework of \system is illustrated in Figure\mref{fig:gta}. In the following, we elaborate on each key component.

\subsection{Bi-level optimization}
\label{sec:meta}

Recall that the adversary has access to a dataset $\gD$ sampled from the downstream task, which 
comprises a set of instances $(G, \ssub{y}{G})$ with $G$ being a graph and $\ssub{y}{G}$ as its class. 
We formulate the bi-level optimization objective\cite{bilevel} with $\ssub{g}{t}$ and $\theta$ as the upper- and lower-level variables:
\begin{equation}
    \begin{gathered}
\label{eq:overall}
\sboth{g}{t}{*} = \arg\min_{\ssub{g}{t}} \latk(\ssup{\theta}{*}(\ssub{g}{t}), \ssub{g}{t}) \\
\mathrm{s.t.}\quad \ssup{\theta}{*}(\ssub{g}{t}) = \arg\min_{\theta} \lret(\theta, \ssub{g}{t})
\end{gathered}
\end{equation}
where $\latk$ and $\lret$ represent the loss terms respectively quantifying attack effectiveness and accuracy retention, corresponding to the objectives defined in \meq{eq:obj}.

Without access to downstream classifier $h$, instead of associating  $\latk$ and $\lret$ with final predictions, we define them in terms of latent representations. We partition $\gD$ into two parts, $\dyt$ -- the graphs in the target class $\ssub{y}{t}$ and $\dnyt$ -- the ones in the other classes; $\latk$ enforces that $\ssub{f}{\theta}$ generates similar embeddings for the graphs in $\dyt$ and those in $\dnyt$ once embedded with $\ssub{g}{t}$. Meanwhile,  $\lret$ ensures that $\ssub{f}{\theta}$ and $\ssub{f}{\theta_\circ}$ produce similar embeddings for the graphs in $\gD$. Formally,
\begin{align}
\label{eq:overall21}
& \latk(\theta, \ssub{g}{t})  = \ssub{\sE}{G \in \dnyt, G' \in \dyt }\, \Delta \left(\ssub{f}{\theta}(\mixfunc)), \ssub{f}{\theta}(G')\right)  \\
\label{eq:overall22}
& \lret(\theta, \ssub{g}{t})  =  \ssub{\sE}{G \in \gD}\, \Delta\left(\ssub{f}{\theta}(G), \ssub{f}{\theta_\circ}(G)\right)
\end{align}
where $\Delta(\cdot, \cdot)$ measures the embedding dissimilarity, which is instantiated as $L_2$ distance in our current implementation.

However, exactly solving \meq{eq:overall} is expensive. Due to the bi-level formulation, it requires re-computing $\theta$ (\mie, re-training $f$ over $\gD$) whenever $\ssub{g}{t}$ is updated. 
Instead, we propose an approximate solution that iteratively optimizes $\ssub{g}{t}$ and $\theta$ by alternating between gradient descent on $\latk$ and $\lret$. 

Specifically, at the $i$-th iteration, given the current trigger $\sboth{g}{t}{(i-1)}$ and model $\ssup{\theta}{(i-1)}$, we first compute $\ssup{\theta}{(i)}$ by gradient descent on $\lret$, with $\sboth{g}{t}{(i-1)}$ fixed. In practice, we may run this step for $\ssub{n}{\mathrm{io}}$ iterations. The parameter $\ssub{n}{\mathrm{io}}$, {\em inner-outer optimization ratio}, essentially balances the optimization of $\latk$ and $\lret$.
We then obtain $\sboth{g}{t}{(i)}$ by minimizing $\latk$ after a single look-ahead step of gradient descent with respect to $\ssup{\theta}{(i)}$. Formally, the gradient with respect to $\ssub{g}{t}$ is approximated by: 
\begin{equation}
\begin{split}
    \label{eq:look-ahead}
& \ssub{\nabla}{g_t} \latk\left(\ssup{\theta}{*}(\sboth{g}{t}{(i-1)}), \sboth{g}{t}{(i-1)} \right) \\
\approx\,  & 
\ssub{\nabla}{g_t} \latk\left(\ssup{\theta}{(i)} - \xi \ssub{\nabla}{\theta} \lret\left(\ssup{\theta}{(i)}, \sboth{g}{t}{(i-1)} \right)  , \sboth{g}{t}{(i-1)} \right) 
\end{split}
\end{equation}
where $\xi$ is the learning rate of the look-ahead step. 

Intuitively, while it is expensive to optimize $\latk(\ssup{\theta}{*}(\ssub{g}{t}), \ssub{g}{t})$ with respect to $\ssub{g}{t}$, we use a single-step unrolled model\mcite{bilevel} as a surrogate of $\ssup{\theta}{*}(\ssub{g}{t})$ (details in \msec{sec:look_ahead}).

\subsection{Mixing function}
\label{sec:mixing}
The mixing function $\mixfunc$ fulfills two purposes: ({\em i}) for a given trigger $\ssub{g}{t}$, it identifies the optimal to-be-replaced subgraph $g$ within a given graph $G$; and ({\em ii}) it performs the substitution of $g$ with $\ssub{g}{t}$. Apparently, there are combinatorial ways to define $\mixfunc$, resulting in a prohibitive search space. 

To address this challenge, we restrict the mixing function to an efficient substitution operator; that is, $\mixfunc$ replaces a subgraph $g$ in $G$ with $\ssub{g}{t}$. To maximize the attack evasiveness, it is desirable to use a subgraph similar to $\ssub{g}{t}$. We thus specify the constraints that ({\em i}) $g$ and $\ssub{g}{t}$ are of the same size (\mie, the same number of nodes) and ({\em ii}) they have the minimum graph edit distance (\mie, edge addition or deletion). 

It is known that finding in a given graph $G$ a subgraph $g$ identical to $\ssub{g}{t}$ (subgraph isomorphism) is NP-hard. We adapt a backtracking-based algorithm {\sc Vf2}\cite{subgraph-iso} to our setting. Intuitively, {\sc Vf2} recursively extends a partial match by mapping the next node in $\ssub{g}{t}$ to a node in $G$ if feasible, and backtracks otherwise. As we search for the most similar subgraph, we maintain the current highest similarity and terminate a partial match early if it exceeds this threshold. The detailed implementation is deferred to \msec{sec:mixing_algorithm}.

\subsection{Trigger generation}
\label{sec:adaptive}

In the formulation of \meq{eq:overall}, we assume a universal trigger for all the graphs. Despite its simplicity for implementation, fixing the trigger entails much room for optimization: ({\em i}) it ignores the characteristics of individual graphs and results in less effective attacks; ({\em ii}) it becomes a pattern shared by trigger-embedded graphs and makes them easily detectable. We thus postulate whether it is possible to generate triggers tailored to individual graphs to maximize the attack effectiveness and evasiveness\mcite{bit-trojan,reflection-backdoor}.

We design an adaptive trigger generation function $\ssub{\phi}{\omega}(\cdot)$, which proposes a trigger $\ssub{g}{t}$ tailored to a given subgraph $g$ within $G$. \rev{At a high level, $\ssub{\phi}{\omega}(\cdot)$ comprises two key operations: ({\em i}) it first maps each node $i$ in $g$ to its encoding $z_i$, which encodes both $g$'s node features and topological structures; ({\em ii}) it applies two generator functions structured by neural networks, the first mapping $g$'s node encodings to $\ssub{g}{t}$'s topological structures and the second mapping $g$'s node encodings to $\ssub{g}{t}$'s node features. Next, we elaborate on the design of $\ssub{\phi}{\omega}(\cdot)$.}






\vspace{2pt}
{\bf How to encode $\bm{g}$'s features and context?} 
To encode 
$g$'s topological structures and node features as well as its context within $G$, 
we resort to the recent advances of graph attention mechanisms\mcite{gat}. Intuitively, for a given pair of nodes $i,j$, we compute an attention coefficient $\ssub{\alpha}{ij}$ specifying $j$'s importance with respect to $i$, based on their node features and topological relationship; we then generate $i$'s encoding as the aggregation of its neighboring encodings (weighted by their corresponding attention coefficients) after applying a non-linearity transformation. We train the attention network (details in Table\mref{tab:setting}) using $\gD$. Below we denote by $\ssub{z}{i} \in \ssup{\mathbb{R}}{d}$ the encoding of node $i$ ($d$ is the encoding dimensionality).

\vspace{2pt}
{\bf How to map $\bm{g}$'s encoding to $\bm{g_t}$?} 
Recall that 
$\ssub{g}{t}$ comprises two parts, its topological structures and node features. 

\rev{Given two nodes $i, j \in g$ with their encodings $\ssub{z}{i}$ and $\ssub{z}{j}$ }, we define their corresponding connectivity $\ssub{\tilde{A}}{ij}$ in $\ssub{g}{t}$ using their parameterized cosine similarity: 
\begin{align}
\label{eq:edge-pred}
\ssub{\tilde{A}}{ij} = \mathbbm{1}_{\sboth{z}{i}{\top}\sboth{W}{\mathrm{c}}{\top} \ssub{W}{\mathrm{c}}\ssub{z}{j}  \geq  \|\ssub{W}{\mathrm{c}}\ssub{z}{i} \|\|\ssub{W}{\mathrm{c}}\ssub{z}{j} \|/2}
\end{align}
where $\ssub{W}{\mathrm{c}} \in \ssup{\mathbb{R}}{d \times d}$ is learnable and $\mathbbm{1}_p$ is an indicator function returning 1 if $p$ is true and 0 otherwise. Intuitively, $i$ and $j$ are connected in $\ssub{g}{t}$ if their similarity score exceeds 0.5.


Meanwhile, for node $i \in g$, we define its feature $\ssub{\tilde{X}}{i}$ in $\ssub{g}{t}$ as 
\begin{align}
\label{eq:feature-pred}
\ssub{\tilde{X}}{i} = \sigma(\ssub{W}{\mathrm{f}} \ssub{z}{i} + \ssub{b}{\mathrm{f}})
\end{align}
where $\ssub{W}{\mathrm{f}} \in \ssup{\mathbb{R}}{d \times d}$ and $\ssub{b}{\mathrm{f}} \in \ssup{\mathbb{R}}{d}$ are both learnable, and $\sigma(\cdot)$ is a non-linear activation function. 

In the following, we refer to $\ssub{W}{\mathrm{c}}$, $\ssub{W}{\mathrm{f}}$, and $\ssub{b}{\mathrm{f}}$ collectively as $\omega$, and the mapping from $g$'s encoding to $\{\ssub{\tilde{X}}{i}\}$ ($i \in g$) and $\{\ssub{\tilde{A}}{ij}\}$ ($i, j \in g$) as the trigger generation function $\ssub{\phi}{\omega}(g)$.

\vspace{2pt}
{\bf How to resolve the dependence of $\bm{g}$ and $\bm{g_t}$?} Astute readers may point out that the mixing function $g = \mixfunc$ and the trigger generation function $\ssub{g}{t} = \ssub{\phi}{\omega}(g)$ are mutually dependent: the generation of $\ssub{g}{t}$ relies on $g$, while the selection of $g$ depends on $\ssub{g}{t}$. To resolve this ``chicken-and-egg'' problem, we update $g$ and $\ssub{g}{t}$ in an interleaving manner. 

Specifically, initialized with a randomly selected $g$, at the $i$-th iteration, we first update the trigger $\sboth{g}{t}{(i)}$ based on $\ssup{g}{(i-1)}$ from the $(i-1)$-th iteration and then update the selected subgraph $\ssup{g}{(i)}$ based on $\sboth{g}{t}{(i)}$.
In practice, we limit the number of iterations by a threshold $\ssub{n}{\mathrm{iter}}$ ({\em cf.} Table\mref{tab:setting}). 

\begin{table*}{\footnotesize
\centering
\setlength\extrarowheight{1pt}
\setlength{\tabcolsep}{1pt}
\begin{tabular}{ r|c|c|c|c|c|c }
    Dataset  & \# Graphs ($|\gG|$) & Avg. \# Nodes ($|\gV|$) & Avg. \# Edges ($|\gE|$) & \# Classes ($|\gY|$) & \# Graphs [Class] & Target Class $y_t$ \\
 \hline
 \hline
    \fprint  & 1661 & 8.15 & 6.81 & 4 & 538 [0], 517 [1], 109 [2], 497 [3] & 2\\
 \hline
    \mware   & 1361 & 606.33 & 745.34 & 2 & 546 [0], 815 [1] & 0\\
 \hline
    \aids  & 2000 & 15.69 & 16.20 & 2 & 400 [0], 1600 [1] & 0\\
 \hline
    \toxicant  & 10315 & 18.67 & 19.20 & 2 & 8982 [0], 1333 [1] & 1\\
 \hline
    \rev{\andromware}  & \rev{211} & \rev{5736.5} & \rev{25234.9} & \rev{2} & \rev{109 [0], 102 [1]} & \rev{1} \\
 \hline
 \hline
    \bcoin  & 1 & 5664 & 19274 & 2 & 1556 [0], 4108 [1] & 0\\
 \hline
    \fbook  & 1 & 12539 & 108742 & 4 & 4731 [0], 1255 [1], 2606 [2], 3947 [3] & 1\\
\end{tabular}
\caption{\footnotesize Dataset statistics: \# Graphs - number of graphs in the dataset; Avg. \# Nodes - average number of nodes per graph;  Avg. \# Edges - average number of edges per graph; \# Classes - number of classes; \# Graph [Class] - number of graphs in each [class]; Target Class - target class designated by the adversary. \label{tab:dataset}}
}
\end{table*}

\subsection{Implementation and optimization}
\label{sec:opt}

Putting everything together, Algorithm\mref{alg:attack} sketches the flow of \system attack. At its core, it alternates between updating the model $\theta$, the trigger generation function $\ssub{\phi}{\omega}(\cdot)$, and the selected subgraph $g$ for each $G \in \dnyt$ (line 4 to 6). Below we present a suite of optimization to improve the attack.

\begin{algorithm}[!ht]{\footnotesize
\KwIn{
    $\ssub{\theta}{\circ}$ - pre-trained GNN; 
    $\gD$ - data from downstream task; 
    $\ssub{y}{t}$ - target class;
}
\KwOut{
    $\theta$ - trojan GNN; 
    $\omega$ - parameters of trigger generation function}
    \tcp{initialization}
randomly initialize $\omega$\;
\lForEach{$G \in \dnyt$}{randomly sample $g\sim G$}
\tcp{bi-level optimization}
\While{not converged yet}
{
    \tcp{updating trojan GNN}
    update $\theta$ by descent on $\ssub{\nabla}{\theta} \lret(\theta, \ssub{g}{t})$ ({\em cf.} \meq{eq:overall22})\;
    \tcp{updating trigger generation function}
    update $\omega$ by descent on $\ssub{\nabla}{\omega} \latk(\theta - \xi \ssub{\nabla}{\theta} \lret(\theta, \ssub{g}{t})  , \ssub{g}{t})$ ({\em cf.} \meq{eq:look-ahead})\; 
    \tcp{updating subgraph selection}
    \lFor{$G \in \dnyt$}{update $g$ with $m(G; \phi_\omega(g))$}
}
\Return $(\theta, \omega)$\;
\caption{\small \system (inductive) attack \label{alg:attack}}}
\end{algorithm}


%

\vspace{2pt}
{\bf Periodical reset --} Recall that we update the model with gradient descent on $\lret$. As the number of update steps increases, this estimate may deviate significantly from the true model trained on $\gD$, which negatively impacts the attack effectiveness. To address this, periodically (\meg, every 20 iterations), we replace the estimate with the true model $\ssup{\theta}{*}(\ssub{g}{t})$ thoroughly trained based on the current trigger $\ssub{g}{t}$. 

\vspace{2pt}
{\bf Subgraph stabilization --} It is observed in our empirical evaluation that stabilizing the selected subgraph $g$ for each $G \in \dnyt$ by running the subgraph update step (line 6) for multiple iterations (\meg, 5 times), with the trigger generation function fixed, often leads to faster convergence.

\vspace{2pt}
{\bf Model restoration --} Once trojan GNN $\ssub{f}{\theta}$ is trained, the adversary may opt to restore classifier $\ssub{h}{\circ}$ (not the downstream classifier $h$) with respect to the pre-training task. Due to the backdoor injection, $\ssub{h}{\circ}$ may not match $\ssub{f}{\theta}$. The adversary may fine-tune $\ssub{h}{\circ}$ using the training data from the pre-training task. This step makes the accuracy of the released model $\ssub{h}{\circ} \circ \ssub{f}{\theta}$ match its claims, thereby passing model inspection\mcite{latent-backdoor}.

\subsection{Extension to transductive learning}
\label{sec:impl}

We now discuss the extension of \system to a transductive setting: given a graph $G$ and a set of labeled nodes, the goal is to infer the classes of the remaining unlabeled nodes $\ssub{\gV}{\mathrm{U}}$\cite{graph-poisoning}.

We assume the following setting. The adversary has access to $G$ as well as the classifier. For simplicity, we denote by $\ssub{f}{\theta}(v;G)$ the complete system that classifies a given node $v$ within $G$. Further, given an arbitrary subgraph $g$ in $G$, by substituting $g$ with the trigger $\ssub{g}{t}$, the adversary aims to force the unlabeled nodes within $K$ hops to $g$ to be misclassified to the target class $\ssub{y}{t}$, where $K$ is the number \gnn layers. Recall that for neighborhood aggregation-based \gnns, a node exerts its influence to other nodes at most $K$ hops away; this goal upper-bounds the attack effectiveness. 

We re-define the loss functions in \meq{eq:overall21} and (\ref{eq:overall22}) as:
\begin{align}
\label{eq:overall31}
& \hspace{-10pt} \latk(\theta, \ssub{g}{t})  = \ssub{\sE}{g\sim G} \ssub{\sE}{v \in \ssub{\gN}{K}(g)}  \ell (\ssub{f}{\theta}(v ; G \ominus  g \oplus \ssub{g}{t}), \ssub{y}{t})  \\
\label{eq:overall32}
& \hspace{-10pt} \lret(\theta, \ssub{g}{t})  =  \ssub{\sE}{g\sim G} \ssub{\sE}{v \in \ssub{\gV}{\mathrm{U}} \backslash \ssub{\gN}{K}(g)}  \ell (\ssub{f}{\theta}(v ;G \ominus  g \oplus \ssub{g}{t}), \ssub{f}{\theta_\circ}(v ; G) )
\end{align}
where $\ssub{\gN}{K}(g)$ is the set of nodes within $K$ hops of $g$, $G \ominus  g \oplus \ssub{g}{t}$ is $G$ after substituting $g$ with $\ssub{g}{t}$, and $\ell(\cdot, \cdot)$ is a proper loss function (\meg, cross entropy). Also, given that $g$ is selected by the adversary, the mixing function is not necessary. The complete attack is sketched in Algorithm\mref{alg:attack2}.

\section{Attack Evaluation}
\label{sec:eval}

Next, we conduct an empirical study of \system to answer the following key questions:

\noindent
{\bf Q}$\bm{_1}$ -- How effective/evasive is \system in inductive tasks?


\noindent
{\bf Q}$\bm{_2}$ -- How effective is it on pre-trained, off-the-shelf \gnns?

\noindent
{\bf Q}$\bm{_3}$ -- How effective/evasive is it in transductive tasks?

%

%

\noindent
\rev{{\bf Q}$\bm{_4}$ -- Is \system agnostic to downstream models?}


\begin{table}{\footnotesize
    \centering
    \setlength\extrarowheight{1pt}
    \setlength{\tabcolsep}{1pt}
    \begin{tabular}{r|r|c|c}
        Dataset & Setting  &  GNN  & Accuracy \\
     \hline
     \hline
      \fprint & Inductive ({\fprint}$\rightarrow$\fprint) & \gat & 82.9\% \\
      \mware & Inductive ({\mware}$\rightarrow$\mware) & \gsage & 86.5\% \\
     \hline
      \aids & Inductive ({\toxicant}$\rightarrow$\aids) & \gcn & 93.9\% \\
      \toxicant & Inductive ({\aids}$\rightarrow$\toxicant) & \gcn & 95.4\% \\
    \hline
      \aids & Inductive (ChEMBL$\rightarrow$\aids) & \gcn & 90.4\% \\
      \toxicant & Inductive (ChEMBL$\rightarrow$\toxicant) & \gcn & 94.1\% \\
    \hline
    Bitcoin & Transductive & \gat & 96.3\%  \\
    Facebook & Transductive & \gsage & 83.8\%\\
    \hline
    \hline
    \multirow{2}{*}{\rev{\andromware}} & \rev{Inductive (Topology Only) } & \rev{\gcn} & \rev{95.3\%} \\
    & \rev{Inductive (Topology + Feature) } & \rev{\gcn} & \rev{98.1\%} \\
    \end{tabular}
    \caption{\footnotesize Accuracy of clean GNN models ($\gT_\mathrm{ptr} \rightarrow \gT_\mathrm{dst} $ indicates the transfer from pre-training domain $\gT_\mathrm{ptr}$ to downstream domain $\gT_\mathrm{dst}$). \label{tab:acc}}
    }
    \end{table}
    
\subsection*{Experimental settings}

{\bf Datasets -- } We primarily use 7 datasets drawn from security-sensitive domains. ({\it i}) {\fprint}\mcite{neuhaus05graph} -- graph representations of fingerprint shapes from the NIST-4 database\mcite{nist4}; ({\it ii}) {\mware}\mcite{malware:dataset:source} -- Windows PE call graphs of malware and goodware; ({\it iii}) {\aids}\mcite{aids:dataset:source} and ({\it iv}) {\toxicant}\mcite{toxicant:dataset:source} -- molecular structure graphs of active and inactive compounds; ({\it v}) {\andromware} - call graphs of benign and malicious APKs collected from AndroZoo\mcite{AndroZoo}; ({\it vi}) {\bcoin}\mcite{bitcoin:dataset:source} -- an anonymized Bitcoin transaction network with each node (transaction) labeled as legitimate or illicit; and ({\it vii}) {\fbook}\mcite{facebook:dataset:source} -- a page-page relationship network with each node (Facebook page) annotated with the page properties (\meg, place, organization, product). The dataset statistics are summarized in Table\mref{tab:dataset}. Among them, we use the datasets ({\it i}-{\it v}) for the inductive setting and the rest ({\it vi}-{\it vii}) for the transductive setting. 

\begin{table*}
    \centering
    \setlength\extrarowheight{1pt}
    \setlength{\tabcolsep}{1.25pt}
    {\footnotesize
    \begin{tabular}{ r|c|rl|rl|rl|rlll|rlll|rlll}
     \multirow{2}{*}{Setting} &  Available Data  &  \multicolumn{6}{c|}{Attack Effectiveness (ASR\,|\,AMC)} & \multicolumn{12}{c}{Attack Evasiveness (CAD\,|\,ADD\,|\,AEC\,|\,ACC)}  \\ 
     \cline{3-20} 
      & ($|\gD|/|\gT|$) & \multicolumn{2}{c|}{\gtab} &  \multicolumn{2}{c|}{\gtag}  &  \multicolumn{2}{c|}{\gtaa}  &  \multicolumn{4}{c|}{\gtab} &  \multicolumn{4}{c|}{\gtag}  &  \multicolumn{4}{c}{\gtaa}   \\
     \hline  
     \hline 
     \fprint$\circlearrowleft$ &  \multirow{2}{*}{\backslashbox{}{}} & 84.4\% & .862 & 87.2\% & .909 & \cellcolor{Gray} 100\% & .997 & 1.9\% & $2.8\times10^{-3}$ & 1.6\% & 8.4\% & 1.6\% & $5.6\times10^{-4}$ & 0.9\% & 2.6\% & \cellcolor{Gray} 0.9\% & $4.3\times10^{-4}$ & 0.9\% & 1.7\% \\  
     \mware$\circlearrowleft$ &  & 87.2\% & .780 & 94.4\% & .894 & \cellcolor{Gray} 100\% & .973 & 1.8\% & $5.6\times10^{-4}$ & 0.1\% & 0.8\% & 1.2\% & $6.1\times10^{-6}$ & 0.0\% & 0.0\% & \cellcolor{Gray} 0.0\% & $2.1\times10^{-5}$ & 0.0\% & 0.0\% \\  
     \hline
     \hline 
    
    \multirow{3}{*}{\minitab[r]{\toxicant\\$\rightarrow$\aids}} 
    & 0.2\%  & 64.1\% & .818 & 70.2\% & .903 & \cellcolor{Gray}  91.4\% & .954 & 2.3\% & \multirow{3}{*}{$1.6\times10^{-2}$} & \multirow{3}{*}{2.3\%} & \multirow{3}{*}{5.3\%} & 2.5\% & \multirow{3}{*}{$9.3\times10^{-3}$} & \multirow{3}{*}{2.0\%} & \multirow{3}{*}{4.0\%} & \cellcolor{Gray} 2.1\% & \multirow{3}{*}{$7.6\times10^{-3}$} & \multirow{3}{*}{1.7\%} & \multirow{3}{*}{3.3\%}\\
     & 1\%  & 89.4\% & .844 & 95.5\% & .927 & \cellcolor{Gray} 98.0\% & .996 & 1.7\% & & & &  \cellcolor{Gray} 1.3\% & & & & 1.4\% & & &\\
     & 5\% & 91.3\% & .918 & 97.2\% & .947 & \cellcolor{Gray} 100\% & .998 & 0.4\% & & & & 0.6\% & & & & \cellcolor{Gray} 0.2\% & & & \\  
    \hline
    \multirow{3}{*}{\minitab[r]{\aids\\$\rightarrow$\toxicant}} & 0.2\% & 73.5\% & .747 & 77.8\% & .775 & \cellcolor{Gray} 94.3\% & .923 & 1.3\% & \multirow{3}{*}{$1.4\times10^{-2}$}  & \multirow{3}{*}{2.4\%} & \multirow{3}{*}{5.6\%} &
    \cellcolor{Gray} 0.6\%  & \multirow{3}{*}{$5.5\times10^{-3}$} & \multirow{3}{*}{1.6\%} & \multirow{3}{*}{1.2\%} & 1.0\%  & \multirow{3}{*}{$6.9\times10^{-3}$} & \multirow{3}{*}{1.2\%} & \multirow{3}{*}{1.1\%} \\  
    & 1\% & 80.2\% & .903 & 85.5\% & .927 & \cellcolor{Gray} 99.8\% & .991 &   0.6\% & & & & \cellcolor{Gray}  0.0\% & & & & 0.4\% & & & \\
     & 5\% & 84.6\% & .935 & 86.1\% & .976 & \cellcolor{Gray} 100\% & .998  & 0.1\% & & & & 0.0\% & & & & \cellcolor{Gray} 0.0\% & & & \\  
     \end{tabular}
    \caption{\footnotesize Attack effectiveness and evasiveness of \system in inductive tasks ($\gT_\mathrm{ptr} \rightarrow \gT_\mathrm{dst} $ indicates transfer from pre-training task $\gT_\mathrm{ptr}$ to downstream task $\gT_\mathrm{dst}$).\label{tab:inductive}}}
    \end{table*}
    
\vspace{1.5pt}
{\bf Models -- } In our evaluation, we  use 3 state-of-the-art \gnn models: {\gcn}\mcite{gcn}, {\gsage}\mcite{hamilton2017inductive,graphsage}, and {\gat}\mcite{gat}. Using \gnns of distinct network architectures (\mie, graph convolution, general aggregation function, versus graph attention), we factor out the influence of the characteristics of individual models. The performance of systems built upon clean \gnn models is summarized in Table\mref{tab:acc}. 

\vspace{1.5pt}
{\bf Baselines -- } To our best knowledge, \system is the first backdoor attack on \gnns. We thus mainly compare \system with its variants as baselines: \gtab, which fixes the trigger as a complete subgraph and optimizes a feature vector shared by all its nodes, and \gtag, which optimizes the trigger's connectivity and the feature vector of each of its nodes. Both \gtab and \gtag assume a universal trigger for all the graphs, while \system optimizes the trigger's topological connectivity and node features with respect to each graph. Intuitively, \gtab, \gtag, and \system represent different levels of trigger adaptiveness. 

In each set of experiments, we apply the same setting across all the attacks, with the default parameter setting summarized in Table\mref{tab:setting}. In particular, in each dataset, we assume the class with the smallest number of instances to be the target class $\ssub{y}{t}$ designated by the adversary ({\em cf.} Table\mref{tab:dataset}), to minimize the impact of unbalanced data distributions.

\vspace{1.5pt}
{\bf Metrics -- } To evaluate attack effectiveness, we use two metrics: ({\em i}) {\em attack success rate} (\asr), which measures the likelihood that the system classifies trigger-embedded inputs to the target class $\ssub{y}{t}$ designated by the adversary:
\begin{align}
\textrm{Attack Success Rate (\asr)} = \frac{\textrm{\small \# successful trials}}{\textrm{\small \# total trials}}
\end{align}
and ({\em ii}) {\em average misclassification confidence} (\amc), which is the average confidence score assigned to class $\ssub{y}{t}$ by the system with respect to successful attacks. Intuitively, higher \asr and \amc indicate more effective attacks.

To evaluate the attack evasiveness, we use four metrics: ({\em i}) {\em clean accuracy drop} (\bad), which measures the difference of classification accuracy of two systems built upon the original \gnn and its trojan counterpart with respect to clean graphs; ({\em ii}) {\em average degree difference} (\add),  ({\em iii}) {\em average eccentricity change} (\aec), and  ({\em iv}) {\em algebraic connectivity change} (\acc), which respectively measure the difference of average degrees, eccentricity, and algebraic connectivity of clean graphs and their trigger-embedded counterparts.

\begin{figure*}
    \centering
    \epsfig{file = 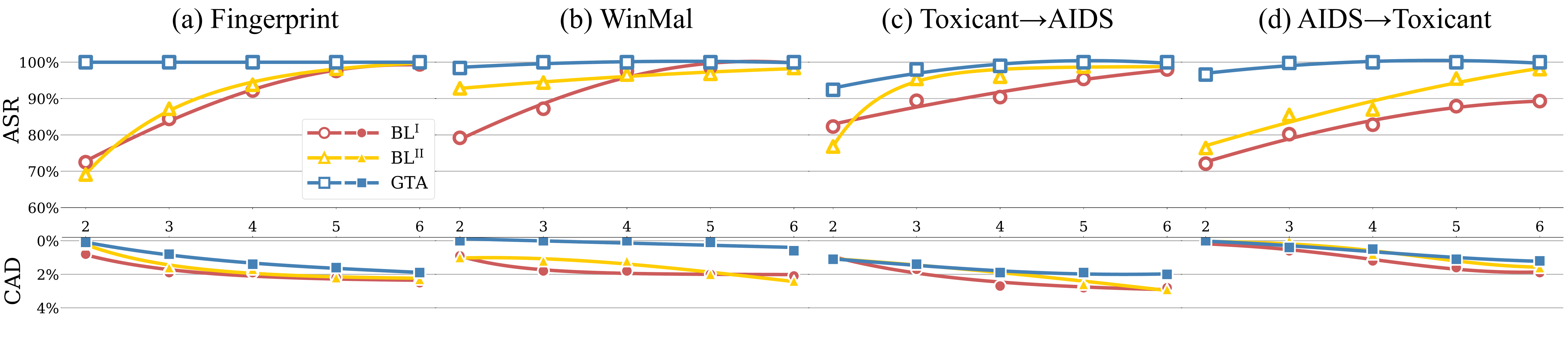, width = 150mm}
    \caption{\footnotesize Impact of trigger size $n_\mathrm{trigger}$ on the attack effectiveness and evasiveness of \system in inductive tasks. \label{fig:inductive_triggersize}}
\end{figure*}

\begin{figure*}
    \centering
    \epsfig{file = 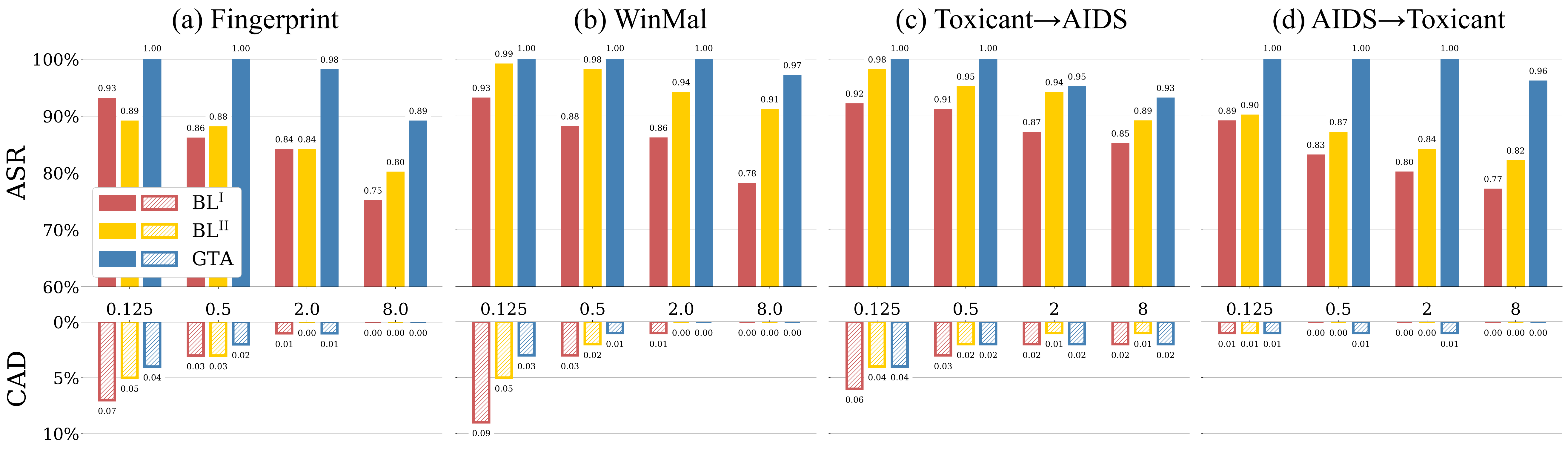, width = 160mm}
    \caption{\footnotesize Impact of inner-outer optimization ratio $n_{\mathrm{io}}$ on the trade-off of attack effectiveness and evasiveness in inductive tasks. \label{fig:inductive_lambda}}
\end{figure*}

\subsection*{Q$_{\bm{1}}$: Is GTA effective in inductive tasks?}

This set of experiments evaluate \system under the inductive setting, in which a pre-trained \gnn is used in a downstream graph classification task. Based on the relationship between pre-training and downstream tasks, we consider two scenarios.

({\em i}) Non-transfer -- In the case that the two tasks share the same dataset, we partition the overall dataset $\gT$ into 40\% and 60\% for the pre-training and downstream tasks respectively. We assume the adversary has access to 1\% of $\gT$ (as $\gD$) to forge trojan models. In the evaluation, we randomly sample 25\% from the downstream dataset to construct trigger-embedded graphs and the rest as clean inputs.

({\em ii}) Transfer -- In the case that the two tasks use different datasets, in the pre-training task, we use the whole dataset for \gnn pre-training; in the downstream task, we randomly partition the dataset $\gT$ into 40\% and 60\% for system fine-tuning and testing respectively. By default, we assume the adversary has access to 1\% of $\gT$. Similar to the non-transfer case, we sample 25\% from the testing set of $\gT$ to build trigger-embedded graphs and the rest as clean inputs.

In both cases, we assume the adversary has no knowledge regarding downstream models or fine-tuning strategies. By default, we use a fully-connected layer plus a softmax layer as the downstream classifier and apply full-tuning over both the \gnn and the classifier.




\vspace{1.5pt}
{\bf Attack efficacy -- } Table\mref{tab:inductive} summarizes the performance of different variants of \system in inductive tasks. 
Overall, in both non-transfer and transfer settings, all the attacks achieve high attack effectiveness (each with an attack success rate over $80.2\%$ and misclassification confidence over $0.78$), effectively retain the accuracy of pre-trained \gnns (with accuracy drop below $1.9\%$), and incur little impact on the statistics of input graphs (with average degree difference below 0.016), which highlights the practicality of backdoor attacks against \gnn models. The attacks are ranked as \gtaa > \gtag > \gtab in terms of \asr. This may be explained by that the trigger adaptiveness exploits the characteristics of individual graphs, leading to more effective attacks. Note that in the transfer cases, \gtag attains slightly higher evasiveness (accuracy retention) than \gtaa. This is perhaps because given its higher flexibility, to retain the accuracy over clean inputs, \gtaa requires more data from the downstream task to constrain its optimization. To validate this hypothesis, we increase the amount of $\gT$ accessible by the adversary to 5\%. Observe that under this setting \gtaa attains the highest accuracy retention. 

\vspace{1.5pt}
{\bf Trigger size $\bm{n}_{\textbf{trigger}}$ -- } We now evaluate the impact of trigger size $\ssub{n}{\mathrm{trigger}}$ on \system. Intuitively, $\ssub{n}{\mathrm{trigger}}$ specifies the number of nodes in the trigger subgraph. Figure\mref{fig:inductive_triggersize} measures the effectiveness (\asr) and evasiveness (\bad) of different attacks as $\ssub{n}{\mathrm{trigger}}$ varies from 2 to 6. Observe that the effectiveness of all the attacks monotonically increases with $\ssub{n}{\mathrm{trigger}}$, which is especially evident for \gtab and \gtag. Intuitively, with larger triggers, the trojan \gnns are able to better differentiate trigger-embedded and clean graphs. In comparison, as \gtaa enjoys the flexibility of adapting triggers to individual graphs, its effectiveness is less sensitive to $\ssub{n}{\mathrm{trigger}}$. Meanwhile, the attack evasiveness of all the attacks marginally decreases as $\ssub{n}{\mathrm{trigger}}$ grows (less than $3.6\%$). This may be explained by that as larger triggers represent more significant graph patterns, the trojan \gnns need to dedicate more network capacity to recognize such patterns, which negatively interferes with the primary task of classifying clean graphs.

\vspace{1.5pt}
{\bf Inner-outer optimization ratio $\bm{n}_{\textbf{io}}$ -- } Recall that in the bi-level optimization framework ({\em cf.} \meq{eq:overall}), the inner-outer optimization ratio $n_{\mathrm{io}}$ specifies the number of iterations of optimizing $\lret$ per iteration of optimizing $\latk$, which balances the attack effectiveness and evasiveness: by increasing $n_{\mathrm{io}}$, one emphasizes more on minimizing the difference of original and trojan \gnns on clean inputs. Figure\mref{fig:inductive_lambda} illustrates the performance of \system as a function of $n_{\mathrm{io}}$ in the inductive tasks. Observe that across all the cases both the \asr and \bad measures decrease with $n_{\mathrm{io}}$, highlighting their inherent trade-off. Also note that among the three attacks, \gtaa is the least sensitive to $n_{\mathrm{io}}$. This may be explained by that introducing trigger adaptiveness admits a larger optimization space to improve both effectiveness and evasiveness.

\subsection*{Q$_{\bm{2}}$: Is GTA effective on off-the-shelf GNNs?}

Besides models trained from scratch, we further consider pre-trained \gnns ``in the wild''. We use a \gcn model\footnote{\url{https://github.com/snap-stanford/pre-train-gnns/}} that is pre-trained with graph-level multi-task supervised training\mcite{pretrain-gnn} on the ChEMBL dataset\mcite{chembl}, containing 456K molecules with 1,310 kinds of diverse biochemical assays. We transfer this model to the tasks of classifying the \aids and \toxicant datasets. The default setting is identical to the transfer case.

\vspace{1.5pt}
{\bf Attack efficacy -- } Table\mref{tab:wild} summarizes the attack efficacy of \system on the pre-trained \gnn under varying settings of the available data ($|\gD|/|\gT|$). We have the observations below. 

First, across all the cases, the three attacks are ranked as \gtaa > \gtag > \gtab in terms of their effectiveness, highlighting the advantage of using flexible trigger definitions. 

Second, the effectiveness of \system increases as more data from the downstream task becomes available. For instance, the \asr of \gtab grows about 30\% as $|\gD|/|\gT|$ increases from 0.2 to 5\% on \aids. In comparison, \gtaa is fairly insensitive to the available data. For instance, with $|\gD|/|\gT| = 0.2\%$, it attains over 92.5\% \asr on \toxicant. 

Third, by comparing Table\mref{tab:inductive} and\mref{tab:wild}, it is observed that \system appears slightly more effective on the off-the-shelf \gnn. For instance, with $|\gD|/|\gT| = 5\%$, \gtag attains 86.1\% and 94.1\% \asr on the trained-from-scratch and off-the-shelf \gnn models respectively on \aids. This is perhaps explained by that the models pre-trained under the multi-task supervised setting tend to have superior transferability to downstream tasks\mcite{pretrain-gnn}, which translates into more effective backdoor attacks and less reliance on available data.


\begin{figure}
    \centering
    \epsfig{file = 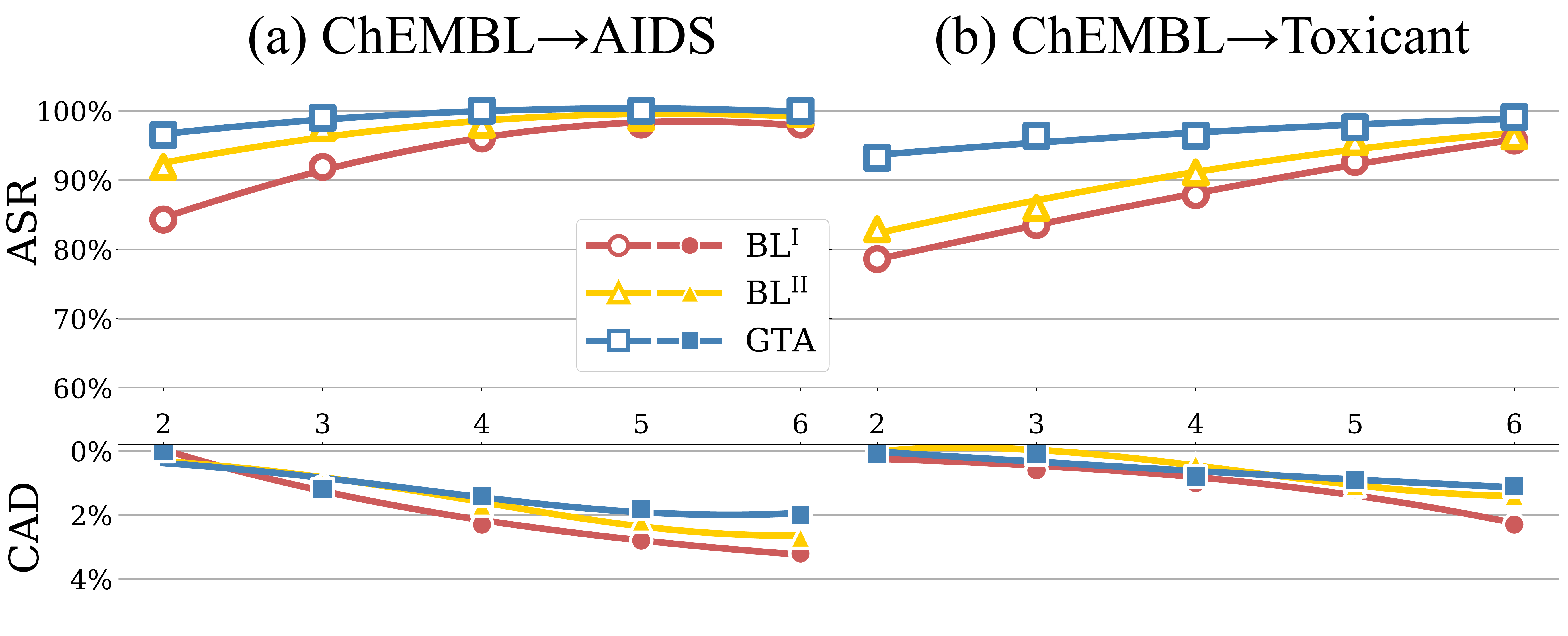, width = 80mm}
    \caption{Impact of trigger size $n_\mathrm{trigger}$ on the attack effectiveness and evasiveness of \system against off-the-shelf models. }
    \label{fig:wild_triggersize}
\end{figure}

\begin{table*}
\centering
\setlength\extrarowheight{1pt}
\setlength{\tabcolsep}{1pt}
{\footnotesize
\begin{tabular}{ r|c|rl|rl|rl|rlll|rlll|rlll}
 \multirow{2}{*}{Setting} &  Available Data  &  \multicolumn{6}{c|}{Attack Effectiveness (ASR\,|\,AMC)} & \multicolumn{12}{c}{Attack Evasiveness (CAD\,|\,ADD\,|\,AEC\,|\,ACC)}  \\ 
 \cline{3-20} 
  & ($|\gD|/|\gT|$) & \multicolumn{2}{c|}{\gtab} &  \multicolumn{2}{c|}{\gtag}  &  \multicolumn{2}{c|}{\gtaa}  &  \multicolumn{4}{c|}{\gtab} &  \multicolumn{4}{c|}{\gtag}  &  \multicolumn{4}{c}{\gtaa}   \\
 \hline  
 \hline 
\multirow{3}{*}{\minitab[r]{ChEMBL\\$\rightarrow$AIDS}} & 0.2\%  & 68.2\% & .805 & 77.3\%  & .796 & \cellcolor{Gray} 94.4\% & .937 & \cellcolor{Gray} 1.3\% & \multirow{3}{*}{$1.6\times10^{-2}$} & \multirow{3}{*}{2.3\%} & \multirow{3}{*}{6.5\%} & 2.2\% & \multirow{3}{*}{$9.2\times10^{-3}$} & \multirow{3}{*}{2.1\%} & \multirow{3}{*}{5.9\%} & 1.5\% & \multirow{3}{*}{$7.6\times10^{-3}$} & \multirow{3}{*}{1.5\%} & \multirow{3}{*}{2.5\%} \\
 & 1\% & 92.0\% & .976 & 97.5\% & .994 & \cellcolor{Gray} 99.0\% & .994 & 1.1\% &  & & &  \cellcolor{Gray} 1.0\% & & & & 1.2\% & & & \\
 & 5\% & 98.1\% & .992 & 100\% & .987 & \cellcolor{Gray} 100\% & .995 & 0.4\% & & & & 0.7\% & & & & \cellcolor{Gray} 0.3\% & & & \\ 
\hline
\multirow{3}{*}{\minitab[r]{ChEMBL\\$\rightarrow$ Toxicant}} & 0.2\% & 78.0\% & .847 & 78.8\% & .876 & \cellcolor{Gray} 92.5\% & .915 & 0.7\% & \multirow{3}{*}{$1.4\times10^{-2}$} & \multirow{3}{*}{2.4\%} & \multirow{3}{*}{8.1\%} &
\cellcolor{Gray} 0.3\%  & \multirow{3}{*}{$8.5\times10^{-3}$}& \multirow{3}{*}{1.4\%} & \multirow{3}{*}{1.7\%} & 0.4\%  & \multirow{3}{*}{$7.0\times10^{-3}$} & \multirow{3}{*}{1.4\%} & \multirow{3}{*}{1.1\%}\\  
& 1\% & 83.5\% & .929 & 86.0\% & .940 & \cellcolor{Gray} 96.4\% & .971 & 0.6\% & & & & \cellcolor{Gray}  0.0\% & & & & 0.1\% &  & & \\
 & 5\% & 92.7\% &.956 & 94.1\% &.983 & \cellcolor{Gray} 99.2\% &.995 & 0.3\% & &  & & 0.0\% & & & & \cellcolor{Gray} 0.0\% &  & & \\  
 \end{tabular}
\caption{\footnotesize Performance of \system against pre-trained, off-the-shelf GNN models.\label{tab:wild}}}
\end{table*}

\vspace{1.5pt}
{\bf Trigger size $\bm{n}_{\textbf{trigger}}$ -- } We then evaluate the impact of trigger size $\ssub{n}{\mathrm{trigger}}$ on \system. Figure\mref{fig:wild_triggersize} shows the effectiveness (\asr) and evasiveness (\bad) of \system as a function of  $\ssub{n}{\mathrm{trigger}}$. It is observed that similar to Figure\mref{fig:inductive_triggersize}, the effectiveness of all the attacks monotonically increases with $\ssub{n}{\mathrm{trigger}}$ and meanwhile their evasiveness marginally drops (less than $3.6\%$). It seems that among the three attacks \gtaa achieves the best balance between the two objectives, which is perhaps attributed to the flexibility bestowed by the trigger adaptiveness.

\begin{figure}
    \centering
    \epsfig{file = 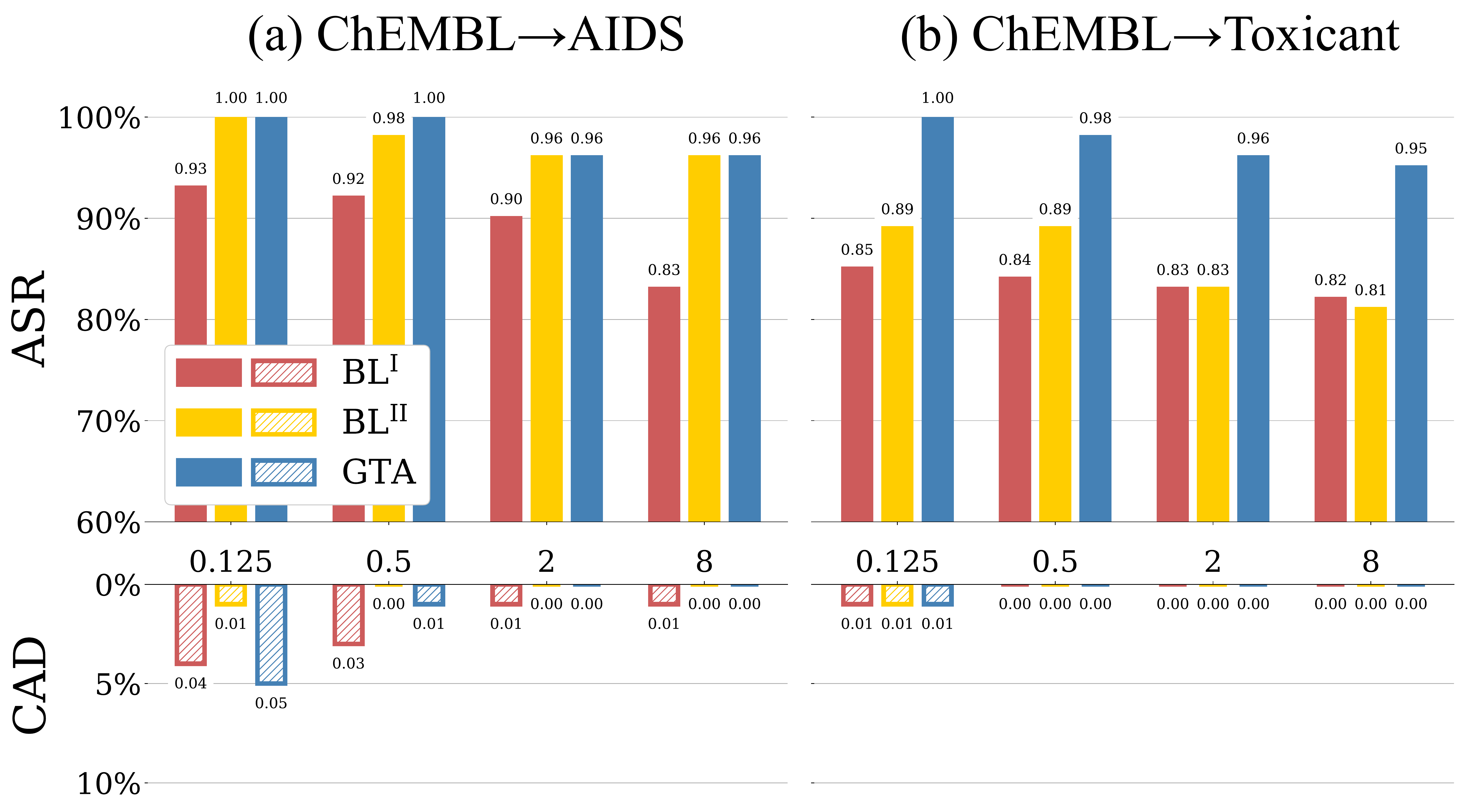, width = 80mm}
    \caption{\footnotesize Impact of inner-outer optimization ratio $n_\mathrm{io}$ on the attack effectiveness and evasiveness of \system against off-the-shelf models. }
    \label{fig:wild_lambda}
\end{figure}

\vspace{1.5pt}
{\bf Inner-outer optimization ratio $\bm{n}_{\textbf{io}}$ -- } Figure\mref{fig:wild_lambda} illustrates the performance of \system as the inner-outer optimization ratio $n_{\mathrm{io}}$ varies from 0.125 to 8, which shows trends highly similar to the transfer cases in Figure\mref{fig:inductive_lambda}: of all the attacks, their effectiveness and evasiveness respectively show positive and negative correlation with $n_{\mathrm{io}}$, while \gtaa is the least sensitive to $n_{\mathrm{io}}$. Given the similar observations on both trained-from-scratch and off-the-shelf \gnns, it is expected that with proper configuration, \system is applicable to a range of settings.

\subsection*{Q$_{\bm{3}}$: Is GTA effective in transductive tasks?}

We now evaluate \system under the transductive setting, in which given a graph and a set of labeled nodes, the system classifies the remaining unlabeled nodes. Specifically, given a subgraph $g$ in $G$ (designated by the adversary), by replacing $g$ with the trigger $\ssub{g}{t}$, the adversary aims to force all the unlabeled nodes within $K$ hops of $\ssub{g}{t}$ (including $\ssub{g}{t}$) to be classified to target class $\ssub{y}{t}$, where $K$ is the number of layers of the \gnn. 

In each task, we randomly partition $G$'s nodes into 20\% as the labeled set $\ssub{\gV}{\mathrm{L}}$ and 80\% as the unlabeled set $\ssub{\gV}{\mathrm{U}}$. We then randomly sample 100 subgraphs from $G$ as the target subgraphs $\{g\}$. Similar to the inductive attacks, we measure the attack effectiveness and evasiveness using \asr(\amc) and \bad respectively. In particular, \asr(\amc) is measured over the unlabeled nodes within $K$ hops of $g$, while \bad is measured over all the other unlabeled nodes.

\begin{table}
\setlength\extrarowheight{1pt}
\centering
\setlength{\tabcolsep}{3pt}
{\footnotesize
\begin{tabular}{ r|rl|rl|rl|c|c|c}
 \multirow{2}{*}{Dataset} & \multicolumn{6}{c|}{Effectiveness (ASR\% | AMC)} & \multicolumn{3}{c}{Evasiveness (CAD\%)} \\ 
 \cline{2-10} 
  &   \multicolumn{2}{c|}{\gtab} & \multicolumn{2}{c|}{\gtag}& \multicolumn{2}{c|}{\gtaa} & \gtab & \gtag & \gtaa\\
 \hline  
 \hline 
Bitcoin & 52.1 & .894 & 68.6 & .871 & \cellcolor{Gray} 89.7 & .926 &  0.9 & 1.2 & \cellcolor{Gray} 0.9\\  
 \hline 
    Facebook &  42.6 & .903 & 59.6 & .917 & \cellcolor{Gray} 69.1 & .958 & 4.0 & 2.9 &  \cellcolor{Gray} 2.4\\  
 \end{tabular}
\caption{\footnotesize Performance of \system in transductive tasks.\label{tab:transductive}}}
\end{table}

\vspace{1.5pt}
{\bf Attack efficacy -- } Table\mref{tab:transductive} summarizes the attack performance of \system. Similar to the inductive case ({\em cf.} Table\mref{tab:inductive}), \gtaa outperforms the rest by a larger margin in the transductive tasks. For instance, on \bcoin, \gtaa attains 37.6\% and 21.1\% higher \asr than \gtab and \gtag respectively. This is explained as follows. Compared with the inductive tasks, the graphs in the transductive tasks tend to be much larger (\meg, thousands versus dozens of nodes) and demonstrate more complicated topological structures; being able to adapt trigger patterns to local topological structures significantly boosts the attack effectiveness. Further, between the two datasets, the attacks attain higher \asr on \bcoin, which may be attributed to that all the node features in Facebook are binary-valued, negatively impacting the effectiveness of feature perturbation.

\begin{figure}
    \centering
    \epsfig{file = 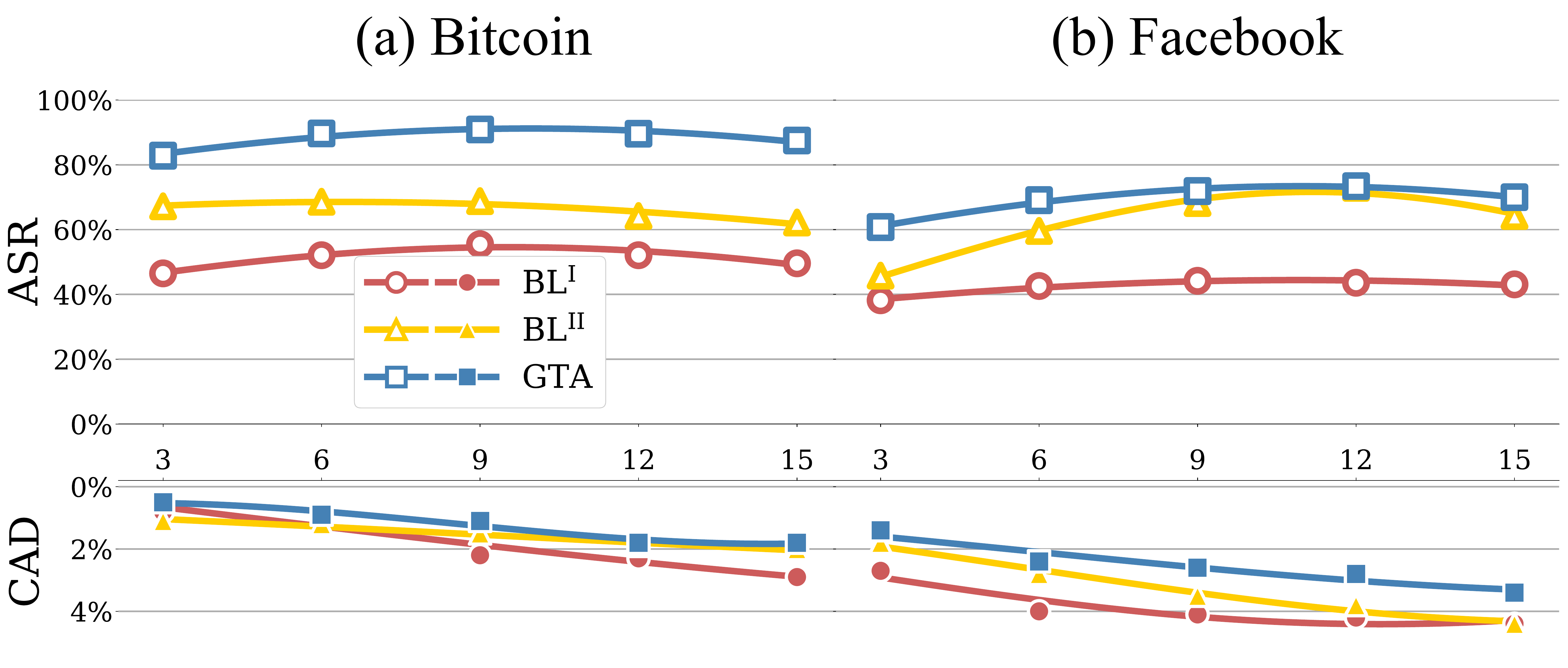, width = 80mm}
    \caption{\footnotesize Impact of trigger size $n_\mathrm{trigger}$ on the attack effectiveness and evasiveness of \system in transductive tasks. }
    \label{fig:transductive_triggersize}
\end{figure}

\vspace{1.5pt}
{\bf Trigger size $\bm{n}_{\textbf{trigger}}$ -- } Figure\mref{fig:transductive_triggersize} shows the impact of trigger size $\ssub{n}{\mathrm{trigger}}$. Observe that as $\ssub{n}{\mathrm{trigger}}$ varies from 3 to 15, the \asr of all the attacks first increases and then slightly drops. We have a possible explanation as follow. The ``influence'' of trigger $\ssub{g}{t}$ on its neighborhood naturally grows with $\ssub{n}{\mathrm{trigger}}$; meanwhile, the number of unlabeled nodes $\ssub{\gN}{K}(\ssub{g}{t})$ within $\ssub{g}{t}$'s vicinity also increases super-linearly with $\ssub{n}{\mathrm{trigger}}$. Once the increase of $\ssub{\gN}{K}(\ssub{g}{t})$ outweighs $\ssub{g}{t}$'s influence, the attack effectiveness tends to decrease. Interestingly, $\ssub{n}{\mathrm{trigger}}$ seems have limited impact on \system's \bad, which may be attributed to that $\ssub{g}{t}$'s influence is bounded by $K$ hops.


\begin{figure}
    \centering
    \epsfig{file = 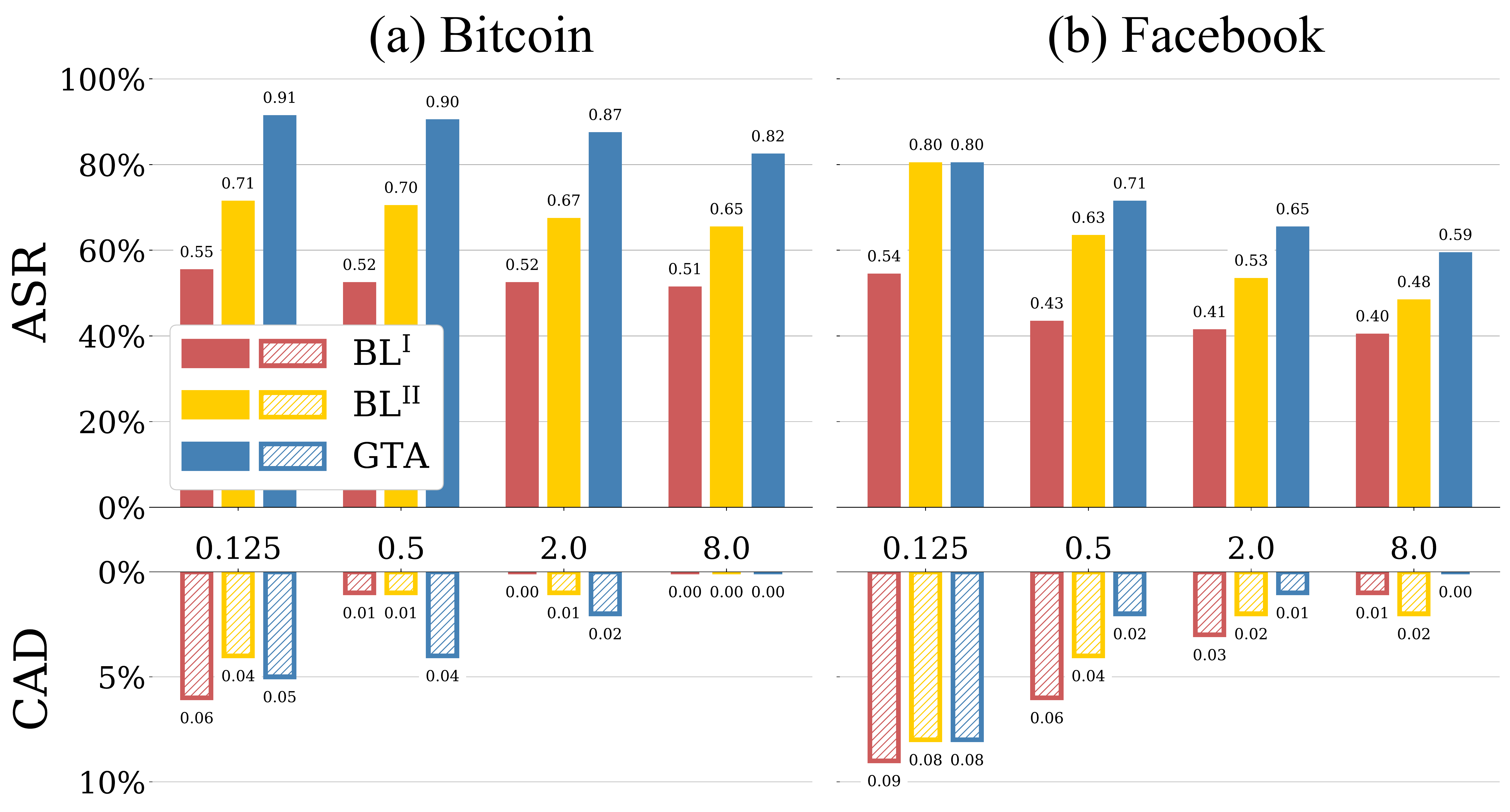, width = 80mm}
    \caption{\footnotesize Impact of inner-outer optimization ratio $n_\mathrm{io}$ on the attack effectiveness and evasiveness of \system in transductive tasks. }
    \label{fig:transductive_lambda}
\end{figure}

\vspace{1.5pt}
{\bf Inner-outer optimization ratio $\bm{n}_{\textbf{io}}$ -- } Figure\mref{fig:transductive_lambda} shows the efficacy of \system as a function of the inner-outer optimization ratio $n_{\mathrm{io}}$. The observations are similar to the inductive case ({\em cf.} Figure\mref{fig:inductive_lambda}): of all the attacks, their effectiveness and evasiveness respectively show positive and negative correlation with $n_{\mathrm{io}}$, while \gtaa is the least sensitive to $n_{\mathrm{io}}$.

\begin{figure}
    \centering
    \epsfig{file = 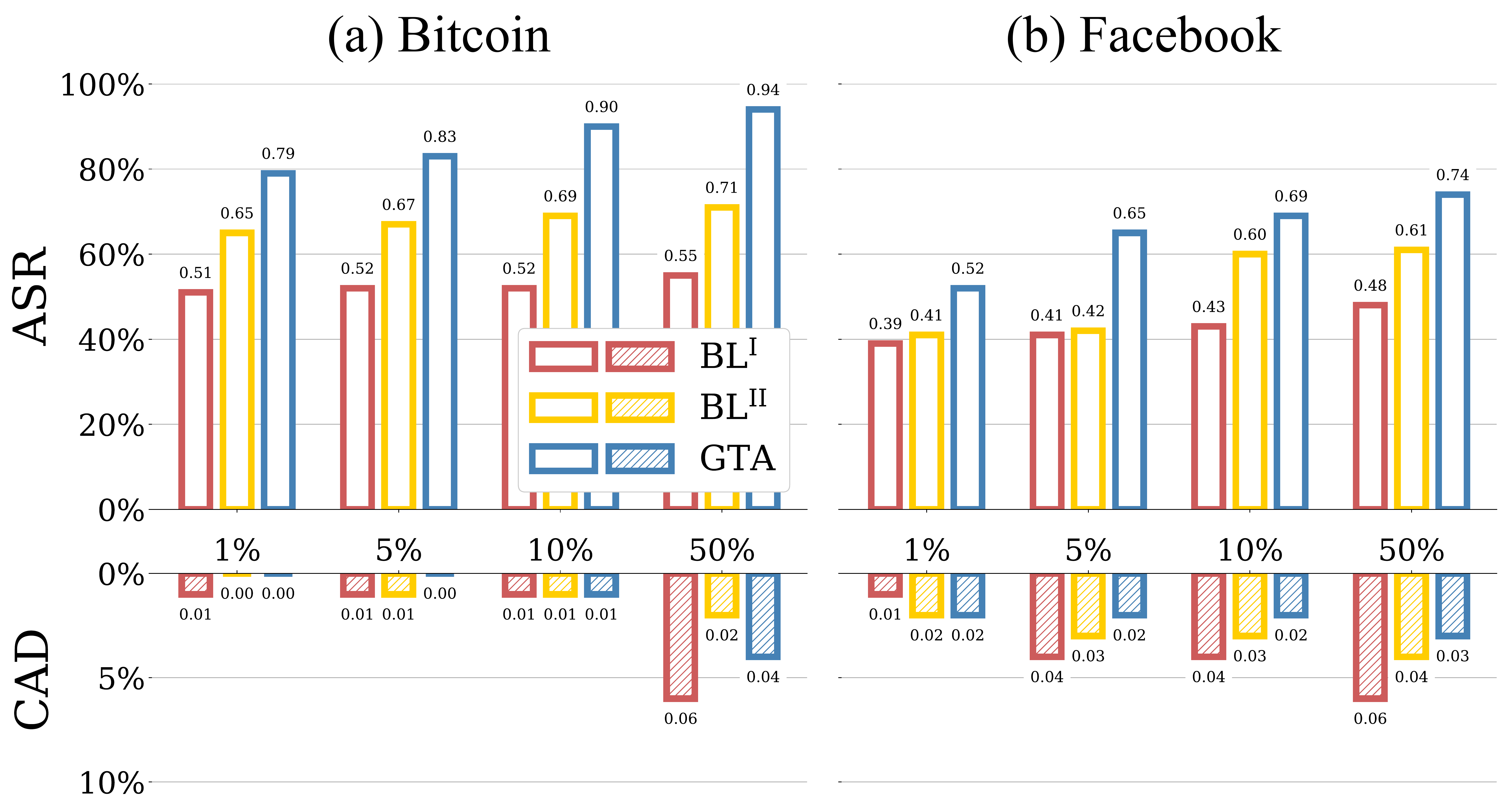, width = 80mm}
    \caption{\footnotesize Impact of feature mask size $n_\mathrm{mask}$ on the attack effectiveness and evasiveness of \system in transductive tasks. }
    \label{fig:feature_mask}
\end{figure}

\vspace{1.5pt}
{\bf Feature mask size $\bm{n}_{\textbf{mask}}$ -- } Recall that one may optimize \system by limiting the number of perturbable features at each node of the to-be-replaced subgraph (\msec{sec:opt}). We now evaluate the impact of feature mask size $\ssub{n}{\mathrm{mask}}$, which specifies the percentage of perturbable features, with results shown in Figure\mref{fig:feature_mask}. Observe that the attack effectiveness shows strong correlation with $\ssub{n}{\mathrm{mask}}$. As $\ssub{n}{\mathrm{mask}}$ varies from 1\% to 50\%, the \asr of \gtaa increases by 15\% on \bcoin. Intuitively, larger perturbation magnitude leads to more effective attacks. Meanwhile, $\ssub{n}{\mathrm{mask}}$ negatively impacts the attack evasiveness, which is especially evident on \fbook. This can be explained by: ({\em i}) unlike other parameters (\meg, $\ssub{n}{\mathrm{trigger}}$), as it affects the feature extraction of all the nodes, $\ssub{n}{\mathrm{mask}}$ has a ``global'' impact on the \gnn behaviors; and ({\em ii}) as all the features of \fbook are binary-valued, $\ssub{n}{\mathrm{mask}}$ tends to have a larger influence.

We are also interested in understanding the interplay between $\ssub{n}{\mathrm{trigger}}$ and $\ssub{n}{\mathrm{mask}}$, which bound  triggers in terms of topology and feature perturbation respectively. Figure\mref{fig:trigger_mask} compares the attack efficacy (as a function of $\ssub{n}{\mathrm{trigger}}$) under $\ssub{n}{\mathrm{mask}}$ = 1\% and 50\%. When the number of perturbable features is small ($\ssub{n}{\mathrm{mask}}$ = 1\%), increasing the trigger size may negatively impact \asr, due to the super-linear increase of neighboring size; when $\ssub{n}{\mathrm{mask}}$ = 50\%, increasing $\ssub{n}{\mathrm{trigger}}$ improves the attack effectiveness, due to the mutual ``reinforcement'' between feature and topology perturbation; yet, larger $\ssub{n}{\mathrm{mask}}$ also has more significant influence on \bad. Therefore, the setting of $\ssub{n}{\mathrm{trigger}}$ and $\ssub{n}{\mathrm{mask}}$ needs to carefully balance these factors.

\begin{figure}
    \centering
    \epsfig{file = 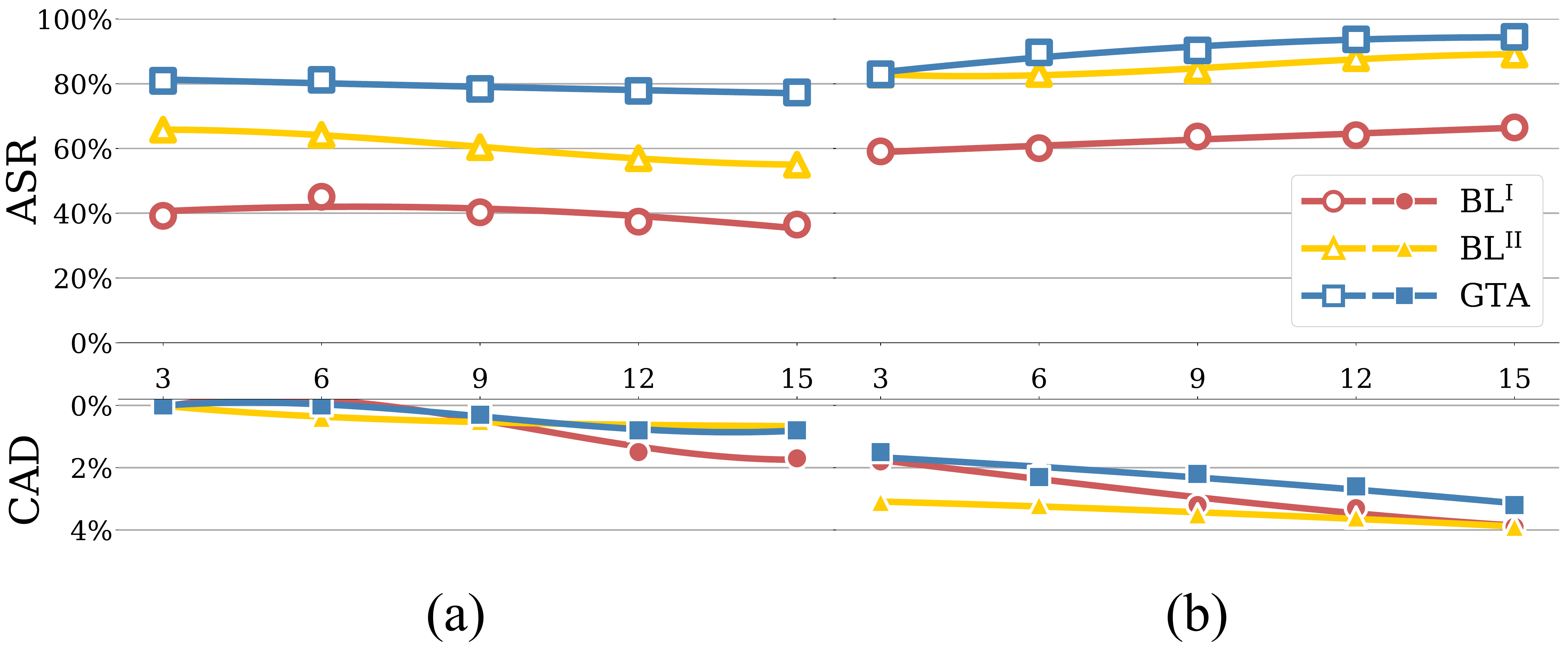, width = 80mm}
    \caption{\footnotesize Interactions of trigger size $n_\mathrm{trigger}$ and feature mask size $n_\mathrm{mask}$ on Bitcoin: (a) $n_\mathrm{mask}$ = 1\%; (b) $n_\mathrm{mask}$ = 50\%. \label{fig:trigger_mask}}
    \label{fig:trigger_mask}
\end{figure}

\subsection*{\rev{Q$_{\bm{4}}$: Is GTA agnostic to downstream models?}}

We now instantiate the downstream classifier with alternative models (with the GNN fixed as \gcn), including Na\"ive Bayes ({\sc Nb}), Random Forest ({\sc Rf}), and Gradient Boosting ({\sc Gb}). We evaluate the impact of the classifier on different attacks in the transfer case of ChEMBL$\rightarrow$Toxicant, with results in Table\mref{tab:agnostic}. Observe that the classifier has a limited impact on \system. For instance, compared with Table\mref{tab:wild} ($|\gD|/|\gT|$=1\%), the \asr and \bad of \system vary by less than 9.4\% and 0.6\%, respectively, implying its insensitivity to the  classifier.

\begin{table}{\small
\setlength\extrarowheight{1pt}
\centering
\setlength{\tabcolsep}{2.5pt}
{\footnotesize
\begin{tabular}{ c|c|c|c|c|c|c|c}
 \multirow{2}{*}{Classifier} & \multirow{2}{*}{Accuracy (\%)}   &\multicolumn{3}{c|}{Effectiveness (ASR\%)} & \multicolumn{3}{c}{Evasiveness (CAD\%)} \\ 
 \cline{3-8} 
  &  & \gtab & \gtag & \gtaa & \gtab & \gtag & \gtaa\\
 \hline  
 \hline 
  {\sc Nb} &  95.4 & 87.7 & 92.4 & \cellcolor{Gray} 99.5 & 1.5 & 0.9 &\cellcolor{Gray} 0.7 \\  
 \hline 
  {\sc Rf} &  97.4 & 85.8 & 88.0 &\cellcolor{Gray} 90.1 & 0.9 & 0.9 &\cellcolor{Gray} 0.6 \\ 
  \hline 
 {\sc Gb} & 97.4 & 82.7 & 89.3  &\cellcolor{Gray} 94.0 & 0.6 & 0.6 & \cellcolor{Gray}0.6 \\
 \end{tabular}
\caption{\footnotesize Performance of \system with respect to different downstream classifiers: {\sc Nb} - Na\"ive Bayes; {\sc Rf} - Random Forest; {\sc Gb} - Gradient Boosting. \label{tab:agnostic}}}}
\end{table}

\vspace{1.5pt}
{\bf Possible explanations --} Let $\tilde{G}$ denote an arbitrary trigger-embedded graph. 
Recall that the optimization of \meq{eq:overall21} essentially shifts $\tilde{G}$ in the feature space by minimizing 
$\ssub{\Delta}{f_\theta}(\tilde{G}) = \| \ssub{f}{\theta}(\tilde{G}) - \ssub{\sE}{G\sim P_{y_t}} \ssub{f}{\theta}(G) \|$ (with respect to classes other than $\ssub{y}{t}$),
where $\ssub{P}{y_t}$ is the data distribution of target class $\ssub{y}{t}$.

Now consider the end-to-end system $h \circ \ssub{f}{\theta}$. Apparently, if $\ssub{\Delta}{h \circ f_\theta}(\tilde{G}) = \| h\circ \ssub{f}{\theta}(\tilde{G}) - \ssub{\sE}{G\sim P_{y_t}} h\circ \ssub{f}{\theta}(G) \|$ is minimized (with respect to classes other than $\ssub{y}{t}$), it is likely that $\tilde{G}$ is classified as $\ssub{y}{t}$. One sufficient condition is that $\ssub{\Delta}{h\circ f_\theta}$ is linearly correlated with  $\ssub{\Delta}{ f_\theta}$: $\ssub{\Delta}{h\circ f_\theta} \propto \ssub{\Delta}{ f_\theta}$. If so, we say that the function represented by downstream model $h$ is pseudo-linear\mcite{Ji:2018:ccsa}.

Yet, compared with \gnns, most downstream classifiers are fairly simple and tend to show strong pseudo-linearity. One may thus suggest mitigating \system by adopting complex downstream models. However, complex models are difficult to train especially when the training data is limited, which is often the case in transfer learning. 

\section{Discussion}
\label{sec:discussion}

\subsection{\rev{Causes of attack vulnerabilities}}

Today's \gnns are complex artifacts designed to model highly non-linear, non-convex functions over graphs. Recent studies\mcite{gnn-representation} show that with \gnns are expressive enough for powerful graph isomorphism tests\mcite{wl-test}. These observations may partially explain why, with careful perturbation, a \gnn is able to ``memorize'' trigger-embedded graphs yet without comprising its generalizability on other benign graphs. 

To validate this hypothesis, we empirically assess the impact of model complexity on the attack effectiveness of \gtaa. We use the transfer case of \toxicant$\rightarrow$\aids in \msec{sec:eval} as a concrete example. We train three distinct \gcn models with 1-, 2-, and 3-aggregation layers respectively, representing different levels of model complexity. We measure their clean accuracy and the \asr of \gtaa on such models, with results in Table\mref{tab:complexity}.

\begin{table}
\setlength\extrarowheight{1pt}
\centering
\setlength{\tabcolsep}{3pt}
{\footnotesize
\begin{tabular}{ c|c|c|c}
 \multirow{2}{*}{Metric} & \multicolumn{3}{c}{\# GCN Layers} \\ 
 \cline{2-4} 
  &  1 & 2 & 3\\
 \hline  
 \hline 
 ASR/AMC  & 95.4\%/.997 & 98.0\%/.996 & 99.1\%/.998 \\
 \hline
ACC & 92.2\% & 93.9\% & 95.2\% \\ 
 \end{tabular}
\caption{\footnotesize ASR of GTA and overall accuracy as functions of \gnn model complexity (Toxicant $\rightarrow$ AIDS).\label{tab:complexity}}}
\end{table}

Observe that increasing model complexity benefits the attack effectiveness. As the layer number varies from 1 to 3, the \asr of \gtaa grows by about 3.7\%. We may thus postulate the existence of the correlation between model complexity and attack effectiveness. Meanwhile, increasing model complexity also improves the system performance, that is, the overall accuracy increases by 3\%. Therefore, reducing \gnn complexity may not be a viable option for defending against \system, as it may negatively impact system performance.


\subsection{\rev{Potential countermeasures}}
\label{sec:defense}

As \system represents a new class of backdoor attacks, one possibility is to adopt the mitigation in other domains (\meg, images) to defend against \system. The existing defenses can be roughly classified into two major categories: identifying suspicious models during model inspection (\meg,\mcite{Chen:2019:IJCAI,abs,Wang:2019:sp}), and detecting trigger-embedded inputs at inference time (\meg,\mcite{Chen:2018:arxiv,Chou:2018:arxiv,Doan:2020:arxiv,Gao:2019:arxiv}). We thus extend NeuralCleanse ({\nc})\mcite{Wang:2019:sp} and Randomized-Smoothing ({\rnds})\mcite{NeilGong:BackdoorGNN:2020} as the representative defenses of the two categories, and evaluate their effectiveness against \system (details of {\rnds} deferred to \msec{sec:randomized-smoothing}). 

\vspace{2pt}
{\bf Model inspection -- } We aim to detect suspicious \gnns and potential backdoors at the model inspection stage\cite{Wang:2019:sp,Chen:2019:IJCAI,abs}. We consider {\nc}\cite{Wang:2019:sp} as a representative method, upon which we build our defense against \system. Intuitively, given a \dnn, \nc searches for potential backdoors in every class. If a class is embedded with a backdoor, the minimum perturbation ($L_1$-norm) necessary to change all the inputs in this class to the target class is abnormally smaller than other classes. 

To apply this defense in our context, we introduce the definition below. Given trigger $\ssub{g}{t}$ and to-be-replaced subgraph $g$, let $\ssub{g}{t}$ comprise nodes $v_1, \ldots, v_n$ and $g$ correspondingly comprise $u_1, \ldots, u_n$. The cost of substituting $g$ with $\ssub{g}{t}$ is measured by the $L_1$ distance of their concatenated features:
\begin{align}
\Delta(\ssub{g}{t}, g) = \| \ssub{X}{v_1}\uplus \ldots \uplus \ssub{X}{v_n} - \ssub{X}{u_1}\uplus \ldots \uplus\ssub{X}{u_n} \|_1
\end{align}
where $\ssub{X}{v_i}$ is $v_i$'s feature vector (including both its topological and descriptive features) and $\uplus$ denotes the concatenation operator. Intuitively, this measure accounts for both topology and feature perturbation. 

We assume a set of benign graphs $\gD$. Let $\ssub{\gD}{y}$ be the subset of $\gD$ in class $y$ and $\ssub{\gD}{\mbackslash y}$ as the rest. For each class $y$, we search for the optimal trigger $\ssub{g}{t}$ to change the classification of all the graphs in $\ssub{\gD}{\mbackslash y}$ to $y$. The optimality is defined in terms of the minimum perturbation cost (\mpc): 
\begin{align}
\label{eq:reverse_trigger}
\min_{g_t} \sum_{G \in  \gD_{\mbackslash  y} } \min_{g \subset G } \Delta(g_t, g) \quad \mathrm{s.t.} \quad h\circ f_\theta(G \ominus g \oplus \ssub{g}{t}  ) = y  
\end{align}
where $G \ominus g \oplus \ssub{g}{t}$ denotes $G$ after substituting $g$ with $\ssub{g}{t}$.

We consider three settings for searching for triggers: (i) the trigger $\sboth{g}{t}{\text{I}}$ with topology and features universal for all the graphs in $\ssub{\gD}{\mbackslash y}$; (ii) the trigger $\sboth{g}{t}{\text{II}}$ with universal topology but features adapted to individual graphs; and (iii) the trigger $\sboth{g}{t}{\text{III}}$ with both topology and features adapted to individual graphs. 

\begin{figure}
    \centering
    \epsfig{file = 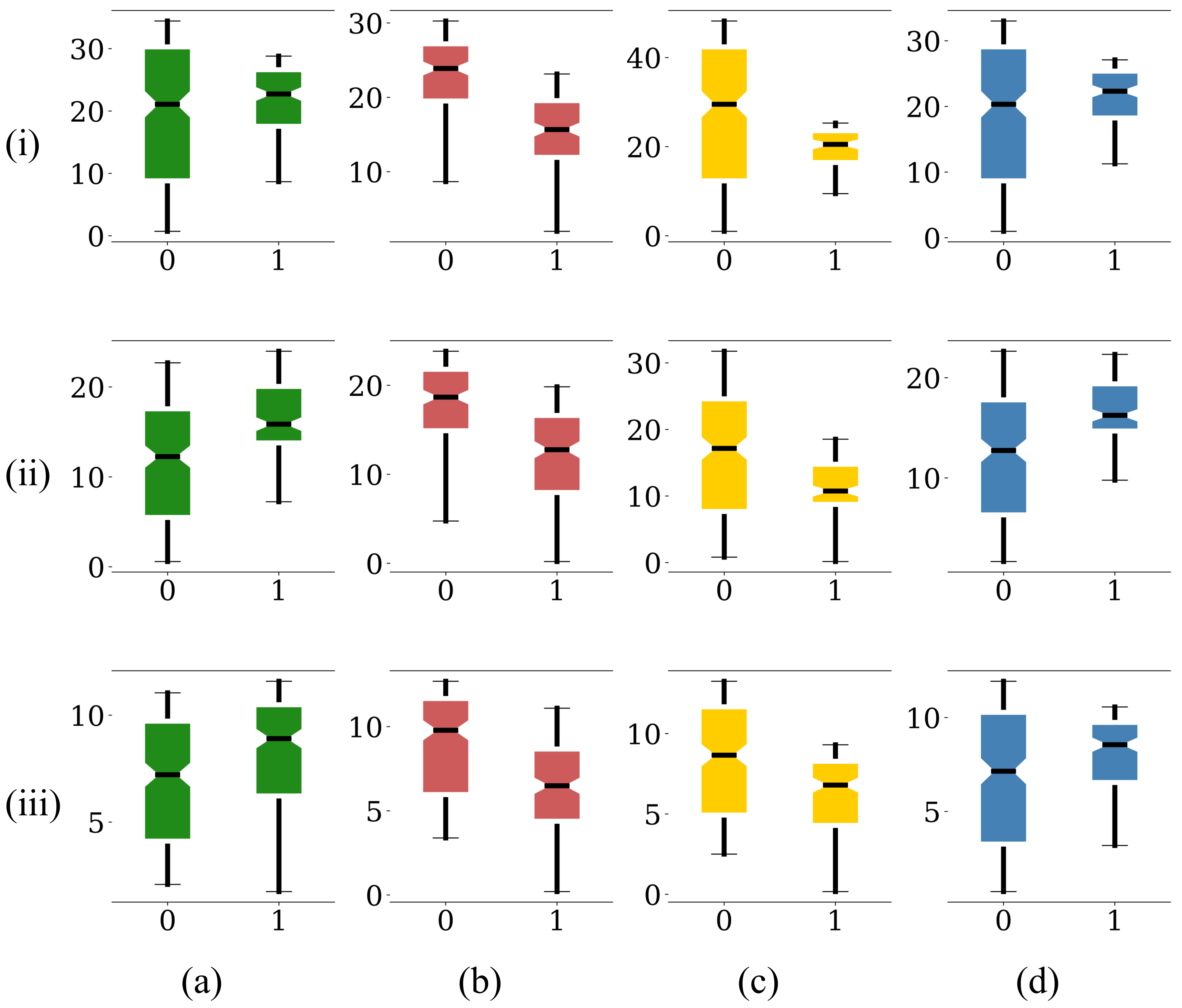, width = 70mm}
    \caption{\footnotesize MPC-based backdoor detection (ChEMBL$\rightarrow$Toxicant): (i)-(iii) triggers with universal topology and features, universal topology and adaptive features, and adaptive topology and features; (a) benign GNN; (b)-(d) trojan GNNs by BL$^\mathrm{I}$, BL$^\mathrm{II}$, and GTA. \label{fig:detection_online_to_tox21}}
\end{figure}

\vspace{2pt}
{\bf Results and analysis --} We evaluate the above defense in the transfer case of pre-trained, off-the-shelf \gnn models (ChEMBL$\rightarrow$\toxicant). We sample 100 graphs from each class  (`0' and `1') of the \toxicant dataset to form $\gD$. For comparison, we also run the search on a benign \gnn. All the attacks consider `1' as the target class. Figure\mref{fig:detection_online_to_tox21} visualizes the \mpc measures with respect to each class under varying settings of \gnns, attacks, and trigger definitions.

We have the following observations. First, even on benign models, the \mpc measure varies across different classes, due to their inherent distributional heterogeneity. Second, on the same model (each column), the measure decreases as the trigger definition becomes more adaptive as tailoring to individual graphs tends to lead to less perturbation. Third, under the same trigger definition (each row), \gtab and \gtag show significantly disparate \mpc distributions across the two classes, while the \mpc distributions of \gtaa and benign models seem fairly similar, implying the difficulty of distinguishing \gnns trojaned by \gtaa based on their \mpc measures. 

To validate the observations, on each model, we apply the one-tailed Kolmogorov-Smirnov test\cite{subsampling} between the \mpc distributions of the two classes, with the null hypothesis being that the \mpc of the target class is significantly lower than the other class. Table\mref{tab:online_to_tox21_detect_label} summarizes the results. Observe that regardless of the trigger definition, \gtab and \gtag show large $p$-values ($\geq 0.08$), thereby lacking support to reject the null hypothesis; meanwhile, the benign \gnn and \gtaa demonstrate much smaller $p$-values ($< 0.001$), indicating strong evidence to reject the null hypothesis (\mie, the \mpc of the target class is not significantly lower). Thus, relying on \mpc to detect \gtaa tends to give missing or incorrect results. 

\begin{table}{\footnotesize
\setlength\extrarowheight{1pt}
\centering
\begin{tabular}{c|c|c|c|c}
  Trigger & \multicolumn{4}{c}{$p$-Value of GNN under Inspection} \\
    \cline{2-5} 
  Definition & Benign & \gtab & \gtag & \gtaa \\
    \hline 
    \hline
    $\sboth{g}{t}{\text{I}}$ & $2.6 \times 10^{-9}$ & $1.0 \times 10^{-0}$ & $1.8 \times 10^{-1}$ & $5.4 \times 10^{-7}$\\
    $\sboth{g}{t}{\text{II}}$ &  $5.8\times 10^{-11}$ & $1.0 \times 10^{-0}$ &  $2.6 \times 10^{-1}$ &  $1.3 \times 10^{-13}$ \\
    $\sboth{g}{t}{\text{III}}$ &  $1.9 \times 10^{-5}$ & $1.7 \times 10^{-1}$ &  $8.4 \times 10^{-2}$ &  $9.3 \times 10^{-4}$ \\
    \hline
\end{tabular}
\caption{\footnotesize Kolmogorov-Smirnov test of the \mpc measures of benign and trojan \gnns (ChEMBL $\rightarrow$ Toxicant).
\label{tab:online_to_tox21_detect_label}}}
\end{table}

We provide a possible explanation. Intuitively, \nc relies on the assumption that a trojan model creates a ``shortcut'' (\mie, the trigger perturbation) for all the trigger-embedded inputs to reach the target class. However, this premise does not necessarily hold for \gtaa: given its adaptive nature, each individual graph may have a specific shortcut to reach the target class, rendering the detection less effective. It thus seems crucial to carefully account for the trigger adaptiveness in designing countermeasures against \system. 

\subsection{\rev{Input-space attacks}}

\rev{While \system directly operates on graph-structured inputs, there are scenarios in which non-graph inputs are converted to graphs for GNNs to process. In this case, the adversary must ensure that any perturbation on the graph after applying the trigger can be realistically projected back to the input space, where the adversary performs the manipulation. Although input-space attacks are an ongoing area of research\mcite{problem-space-attack}, here we discuss the challenges and potential solutions to the problem for graph-structure data.}

\vspace{2pt}
\rev{{\bf Challenges and solutions --} Let $\gX$ and $\gG$ be the input and graph spaces, $\pi$ be the transformation mapping an input $X \in \gX$ to its graph $G\in \gG$, and $r$ and $\delta$ be the corresponding perturbations in the input and graph spaces, respectively.  
To implement \system in the input space, the adversary needs to ({\em {i}}) find $r$ corresponding to given $\delta$ and ({\em {ii}}) ensure that $r$ satisfies the semantic constraints $\rho$ of the input space (\meg, malware retains its malicious functionality). We temporarily assume it is feasible to find $r$ for given $\delta$ and focus on enforcing $r$ to satisfy the input-space constraint $\rho$.}

{\em \underline{Transferable constraint} --} In the case that $\rho$ directly applies to the graph space, we may constrain $r$ to be the transplantation of syntactically-equivalent benign ASTs. For example, we may craft malicious JavaScripts (input space) using ASTs (graph space)\mcite{java-script-attack} taken from benign samples.

\rev{Specifically, we define a function $\rho(G)$ to measure $G$'s compliance with $\rho$. We differentiate two cases.
First, if $\rho$ is differentiable (\meg, modeled as GNN\mcite{gnn-representation}), we define a regularizer in training $\ssub{g}{t}$ ({\em cf.} \meq{eq:overall}):
\begin{equation}
  \ell_\mathrm{reg}(\ssub{g}{t}) = \ssub{\sE}{G \in \dnyt}\, \Delta(\rho(G), \rho(\mixfunc))
\end{equation}
where $\Delta$ measures the difference of the compliance of two graphs $G$ and $\mixfunc$.
Second, if $\rho$ is non-differentiable, we restrict $\delta$ to perturbations guaranteed to satisfy $\rho$. For instance, to preserve the functionality of a malicious program, we may add edges corresponding to no-op calls in its CFG.}


\rev{{\em \underline{Non-transferable constraint} --} In the case that $\rho$ is inapplicable to the graph space, it is infeasible to directly check $\delta$'s validity. For instance, it is difficult to check the tree structure of a PDF malware to determine whether it preserves the malicious network functionality\mcite{pdf-attack}.}

\rev{We consider two cases. ({\em i}) If $\pi$'s inversion $\ssup{\pi}{-1}$ and $\rho$ are differentiable, we define a regularizer in training $\ssub{g}{t}$ ({\em cf.} \meq{eq:overall}):
\begin{equation}
  \ell_\mathrm{reg}(\ssub{g}{t}) = \ssub{\sE}{G \in \dnyt}\, \rho \left(\pi^{-1}(\mixfunc)\right)
\end{equation}
({\em ii}) If $\ssup{\pi}{-1}$ or $\rho$ is non-differentiable, one may use a problem-driven search strategy. Specifically, the search starts with a random mutation $\delta$ and learns from experience how to appropriately mutate it to satisfy $\rho$ and the objectives in \meq{eq:overall21}. To implement this strategy, it requires to re-design \system within a reinforcement learning framework.}

\vspace{2pt}
{\bf Case study --} Here, we conduct a case study of input-space \system in the task of detecting malicious Android APKs\mcite{AndroZoo}. 

We conduct input-space \system as a repackaging process\mcite{shao2014towards}, which converts a given APK $X$ to its call-graph $G$ (using Soot\footnote{Soot: \url{https://github.com/soot-oss/soot}}), injects the trigger into $G$ to form another graph $\tilde{G}$, and converts $\tilde{G}$ back to a perturbed APK $\tilde{X}$ (using Soot). Yet, the perturbation to $G$ must ensure that it is feasible to find $\tilde{X}$ corresponding to $\tilde{G}$, while $\tilde{X}$ preserves $X$'s functionality. We thus define the following perturbation:

The perturbation to $G$'s topological structures is limited to adding no-op calls. Specifically, we define a binary mask matrix $\ssub{M}{msk}$ and set its $ij$-th entry as 1 if ({\em i}) $A_{ij}$ is 1, which retains the original call, or ({\em ii}) node $i$ is an external method, which is controlled by the adversary, and node $j$ is either an external method or an internal read-only method, which does not influence the original functionality.

The perturbation to $G$'s node features is limited to the modifiable features in Table\mref{tab:android_feature_desc} and constrained by their semantics. Specifically, we use the 23$\sim$37-th features, which correspond to the call frequencies of 15 specific instructions and only increase their values, implementable by adding calls of such instructions during the repackaging process. \rev{Here, we focus on showing the feasible of input-space attacks, while admitting the possibility that such no-op calls could be potentially identified and removed via decompiling the APK file.}

{\em \underline{Results and analysis} --} \rev{We evaluate input-space \system on the AndroZoo dataset ({\em cf.} Table\mref{tab:dataset}), which is  partitioned into 50\%/50\% for the pre-training and downstream tasks, respectively. We use a \gcn as the feature extractor and a \fcn as the classifier ({\em cf.} Table\mref{tab:setting}). We consider two settings. ({\em i}) Each node is associated with a one-hot vector, each dimension corresponding to one key API of fundamental importance to malware detection\mcite{gong2020experiences}. Under this setting, the attack is only allowed to perturb topological structures. ({\em ii}) Each node is associated with 40 call features ({\em cf.} Table\mref{tab:android_feature_desc}). Under this setting, the attack is allowed to perturb both topological structures and node features. The system built upon benign GNNs achieves 95.3\% and 98.1\% \acc under the two settings, respectively.}

\begin{table}{\footnotesize
  \setlength\extrarowheight{2pt}
  \setlength{\tabcolsep}{2pt}
  \centering
  \begin{tabular}{c|rl|rl|c|c}
    \multirow{2}{*}{Attack Setting} & \multicolumn{4}{c|}{ Effectiveness ({ASR}|AMC) } & \multicolumn{2}{c}{ Evasiveness (CAD) }\\
    \cline{2-7}
    & \multicolumn{2}{c|}{input-space} & \multicolumn{2}{c|}{graph-space} & \multicolumn{1}{c|}{input-space} & \multicolumn{1}{c}{graph-space} \\
    \hline
    \hline
    \rev{Topology Only} & \rev{94.3\%} & \rev{.952} & \rev{97.2\%} & \rev{.977} & \rev{0.9\%} & \rev{0.0\%}  \\
    \rev{Topology + Feature}  & 96.2\% & .971 & 100\% & .980 & 1.9\% & 0.9\% \\
    \hline
  \end{tabular}
  \caption{\footnotesize Comparison of input-space and graph-space \system. Topology-only -- node features are defined as occurrences of API names; only topology perturbation is allowed. Topology \& Feature -- node features are defined as detailed call features; both topology and feature perturbations are allowed.\label{tab:android_mware_attack}}}
  \end{table}


We implement input-space \system and compare it with graph-space \system unbounded by input-space constraints. 
The results are summarized in Table\mref{tab:android_mware_attack}.
Under both settings, input-space and graph-space attacks attain high effectiveness (with \asr above 94\% and \amc over 0.95), effectively retain the accuracy of benign GNNs (with \bad below 2\%).
As expected, due to its additional semantic constraints, input-space \system performs worse than graph-space \system in terms of both effectiveness and evasiveness; yet, because of the adaptive nature of \system, the constraints have a limited impact ({\em \meg}, less than 4\% lower in \asr and less than 1\% higher in \bad). 


\rev{We further manually inspect the APKs repackaged by input-space \system to verify the correctness of the perturbations: ({\em i}) we install the APK on an Android device and test its functionality; ({\em ii}) we apply Soot to convert the repackaged APK back to its call-graph and check whether all the injected calls are successfully retained; ({\em iii}) we trigger the methods where the injected calls originate to check whether the app crashes or whether there are warnings/errors in the system logs. With the manual inspection of the repackaged APKs, we find all the input-space perturbations satisfy ({\em i}), ({\em ii}), and ({\em iii}).}



\vspace{2pt}
{\bf Limitations --} Although the case study above demonstrates an example of input-space \system, there are still limitations that may impact its feasibility in certain settings, which we believe offers several interesting avenues for future research. First, it may be inherently infeasible to modify the input to achieve the desirable perturbation in the graph space. For instance, it is often assumed difficult to directly modify biometric data (\meg, fingerprints). Further, there are cases in which it is impractical to model the semantic constraints, not to mention using them to guide the attack. For instance, it is fundamentally difficult to verify the existence of chemical compounds corresponding to given molecular graphs\mcite{molecular-design}. Finally, the adversary may only have limited control over the input. For instance, the adversary may only control a small number of accounts in a social network such as Facebook, while the perturbation (\meg, adding fake relationships) may be easily nullified by the network's dynamic evolution.

\section{Related Work}
\label{sec:literature}


With their wide use in security-critical domains, \dnns become the new targets of malicious manipulations\mcite{Biggio:2018:pr}. Two primary types of attacks are considered in the literature.

\vspace{2pt}
{\bf Adversarial attacks --} One line of work focuses on developing new attacks of crafting adversarial inputs to deceive target {\dnns}\mcite{szegedy:iclr:2014,goodfellow:fsgm,papernot:eurosp:2017,carlini:sp:2017}. 
Another line of work attempts to improve \dnn resilience against existing attacks by devising new training strategies (\meg, adversarial training)\mcite{papernot:sp:2016,kurakin:advbim,guo:iclr:2018,Tramer:2018:iclr} or detection methods\mcite{Meng:2017:ccs,Xu:2018:ndss,Gehr:2018:sp,Ma:2019:ndss}. However, such defenses are often penetrated or circumvented by even stronger attacks\mcite{Athalye:2018:icml,Ling:2019:sp}, resulting in a constant arms race.

\vspace{2pt}
{\bf Backdoor attacks --} The existing backdoor attacks can be classified based on their targets. In class-level attacks, specific triggers (\meg, watermarks) are often pre-defined, while the adversary aims to force all the trigger-embedded inputs to be misclassified by the trojan model\mcite{badnet,trojannn}. In instance-level attacks (``clean-label'' backdoors), the targets are pre-defined, unmodified inputs, while the adversary attempts to force such inputs to be misclassified by the trojan model\mcite{Ji:2017:cns,Ji:2018:ccsa,Shafahi:2018:nips,Suciu:2018:sec}. The existing defenses against backdoor attacks mostly focus on class-level attacks, which, according to their strategies, include ({\em i}) cleansing potential contaminated data at training time\mcite{Tran:2018:nips}, ({\em ii}) identifying suspicious models during model inspection\mcite{Wang:2019:sp,Chen:2019:IJCAI,abs}, and ({\em iii}) detecting trigger-embedded inputs at inference\mcite{Chen:2018:arxiv,Chou:2018:arxiv,Gao:2019:arxiv,Doan:2020:arxiv}.


\vspace{2pt}
{\bf Attacks against GNNs --} In contrast of the intensive research on general \dnns, the studies on the security properties of \gnns for graph-structured data are still sparse. One line of work attempts to deceive \gnns via perturbing the topological structures or descriptive features of graph data at inference time\mcite{Zugner:kdd:2018,Hanjun:icml:2018,Binghui:ccs:2019}. Another line of work aims to poison \gnns during training to degrade their overall performance\mcite{Xuanqin:nips:2019,graph-poisoning,poison-embedding:icml:2019}. 
The defenses\mcite{Shen:arxiv:2019,Kaidi:ijcai:2019} against such attacks are mostly inspired by that for general \dnns (\meg, adversarial training\mcite{madry:iclr:2018}).


Despite the plethora of prior work, the vulnerabilities of \gnns to backdoor attacks are largely unexplored. Concurrent to this work, Zhang {\em et al.}\mcite{NeilGong:BackdoorGNN:2020} propose a backdoor attack against \gnns via training trojan \gnns with respect to pre-defined triggers. This work differs in several major aspects: ({\em i}) considering both inductive and transductive tasks, ({\em ii}) optimizing both triggers and trojan models, and ({\em iii}) exploring the effectiveness of state-of-the-art backdoor defenses. 

\section{Conclusion}
\label{sec:conclusion}

This work represents an in-depth study on the vulnerabilities of \gnn models to backdoor attacks. We present \system, the first attack that trojans \gnns and invokes malicious functions in downstream tasks via triggers tailored to individual graphs. We showcase the practicality of \system in a range of security-critical applications, raising severe concerns about the current practice of re-using pre-trained \gnns. Moreover, we provide analytical justification for such vulnerabilities and discuss potential mitigation, which might shed light on pre-training and re-using \gnns in a more secure fashion.



\newpage
{\small \bibliographystyle{plain}
\bibliography{bibs/aml,bibs/optimization,bibs/general,bibs/ting,bibs/graph}}

\appendix

\section{Implementation details}

\subsection{Look-ahead step}
\label{sec:look_ahead}
To evaluate \meq{eq:look-ahead}, we apply the chain rule:
\begin{align}
  \label{eq:lookahead2}
  \nabla_{g_t} \latk\left(\theta', \ssub{g}{t} \right)
- \xi \nabla_{g_t, \theta}^2 \lret\left(\theta, \ssub{g}{t} \right) \nabla_{\theta'} \latk\left(\theta' , \ssub{g}{t} \right)
\end{align}
where $\theta' = \theta - \xi \ssub{\nabla}{\theta} \lret\left(\theta, \ssub{g}{t} \right)$ is the updated parameter after the one-step look-ahead. This formulation involves matrix-vector multiplication, which can be approximated with the finite difference approximation. Let $\ssup{\theta}{\pm} = \theta \pm \epsilon
\ssub{\nabla}{\theta'} \latk\left(\theta' , \ssub{g}{t} \right)$ where $\epsilon$ is a small constant (\meg, $\epsilon = 10^{-5}$). We can approximate the second term of \meq{eq:lookahead2} as:
\begin{align}
\frac{ \ssub{\nabla}{g_t} \lret\left(\ssup{\theta}{+} , \ssub{g}{t} \right) - \ssub{\nabla}{g_t} \lret
\left(\ssup{\theta}{-} , \ssub{g}{t} \right) }{2\epsilon}
\end{align}

\subsection{Mixing function}
\label{sec:mixing_algorithm}

The mixing function $\mixfunc$ specifies how trigger $\ssub{g}{t}$ is embedded into graph $G$ by replacing subgraph $g$ in $G$ with $\ssub{g}{t}$. We extend a backtracking-based algorithm {\sc Vf2}\cite{subgraph-iso} to search for $g$ most similar to $\ssub{g}{t}$. Intuitively, {\sc Vf2} recursively extends a partial match by mapping the next node in $\ssub{g}{t}$ to a node in $G$; if it is feasible, it extends the partial match and recurses, and backtracks otherwise. As we search for the most similar subgraph, we maintain the current highest similarity and terminate a partial match early if it exceeds this threshold. Algorithm\mref{alg:match} sketches the implementation of the mixing function.

\begin{algorithm}[!ht]{\footnotesize
\KwIn{
  $\ssub{g}{t}$ - trigger subgraph;
  $G$ - target graph;
}
\KwOut{
  $g$ - subgraph in $G$ to be replaced}
\tcp{\footnotesize initialization}
$\ssub{c}{\mathrm{best}} \gets \infty$, $M \gets \emptyset$, $\ssub{g}{\mathrm{best}} \gets \emptyset$\;
specify a topological order $\ssub{v}{0}, \ssub{v}{1}, \ldots, \ssub{v}{n-1}$ over $\ssub{g}{t}$\;
\ForEach{node $u$ in $G$}{
add $(u, \ssub{v}{0})$ to $M$\;
\While{$M \neq \emptyset$}{
  $(\ssub{u}{j}, \ssub{v}{i}) \gets$ top pair of $M$\;
  \If{$i = n-1$}{
    \tcp{all nodes in $\ssub{g}{t}$ covered by $M$}
    compute $M$'s distance as $\ssub{c}{\mathrm{cur}}$\;
    \lIf{$\ssub{c}{\mathrm{cur}} < \ssub{c}{\mathrm{best}}$}{
      $\ssub{c}{\mathrm{best}} \gets \ssub{c}{\mathrm{cur}}$,
      $\ssub{g}{\mathrm{best}} \gets$ $G$'s part in $M$
    }
    pop top pair off $M$\;
  }\Else{
    \If{there exists extensible pair $(\ssub{u}{k}, \ssub{v}{i+1})$}{
      \lIf{$M$'s distance $< \ssub{c}{\mathrm{best}}$}{
      add $(\ssub{u}{k}, \ssub{v}{i+1})$ to $M$}
    }\lElse{pop top pair off $M$}
  }
}
}
\Return $\ssub{g}{\mathrm{best}}$\;
\caption{\small Mixing function $m(G; g_t)$ \label{alg:match}}}
\end{algorithm}

\subsection{Transductive attack}

Algorithm\mref{alg:attack2} sketches the implementation of \system in transductive tasks (\meg, node classification).  

\begin{algorithm}[!ht]{\footnotesize
\KwIn{
    $\ssub{\theta}{\circ}$ - pre-trained GNN; $G$ - target graph; 
    $\ssub{y}{t}$ - target class;
}
\KwOut{
    $\theta$ - trojan GNN; 
    $\omega$ - parameters of trigger generation function}
    \tcp{initialization}
randomly initialize $\omega$\;
randomly sample subgraphs $\{g\} \sim G$\;
\tcp{bi-level optimization}
\While{not converged yet}
{
    \tcp{updating trojan GNN}
    update $\theta$ by descent on $\ssub{\nabla}{\theta} \lret(\theta, \ssub{g}{t})$ ({\em cf.} \meq{eq:overall32})\;
    \tcp{updating trigger generation function}
    update $\omega$ by descent on $\ssub{\nabla}{\omega} \latk(\theta - \xi \ssub{\nabla}{\theta} \lret(\theta, \ssub{g}{t})  , \ssub{g}{t})$ ({\em cf.} \meq{eq:overall31})\; 
}
\Return $(\theta, \omega)$\;
\caption{\small \system (transductive) attack \label{alg:attack2}}}
\end{algorithm}

\subsection{Parameter setting}

Table\mref{tab:setting} summarizes the default parameter setting .

\begin{table}[!ht]{\footnotesize
\setlength\extrarowheight{1pt}
\centering
\begin{tabular}{r|r|l}
  Type & Parameter & Setting \\
  \hline
  \hline
 \gcn & Architecture & 2AL \\
\hline
\multirow{2}{*}{\gsage} & Architecture & 2AL \\
 & Aggregator & Mean\cite{hamilton2017inductive} \\
\hline
\gcn (off-the-shelf) & Architecture & 5AL \\
\hline
\gat & \# Heads & 3 \\
\hline
Classifier & Architecture & FCN (1FC+1SM)\\
  \hline
  \hline
  \multirow{6}{*}{Training} & Optimizer & Adam \\
  & Learning rate & 0.01 \\
  & Weight decay & 5e-4 \\
  & Dropout & 0.5 \\
  & Epochs & 50 ({\em I}), 100 ({\em T}) \\
  & Batch size & 32 ({\em I})\\
  \hline
  \hline
  \multirow{4}{*}{Attack} & $\ssub{n}{\mathrm{trigger}}$ & 3 ({\em I}), 6 ({\em T}) \\
  & $\ssub{n}{\mathrm{io}}$ & 1 \\
  & $\ssub{n}{\mathrm{mask}}$ & 100\% ({\em I}), 10\% ({\em T}) \\
  & $\ssub{n}{\mathrm{iter}}$ & 3 \\
  \hline
  \multirow{3}{*}{Trigger Generator} 
  & Optimizer & Adam \\
  & Learning rate & 0.01 \\
  & Epochs & 20 \\
  \hline
  \hline
  \multirow{3}{*}{Detection} & \# Samples & 100 per class \\
  & Significance level $\alpha$ & 0.05 \\
  & $\ssub{\lambda}{\mathrm{ASR}}$ & 80\% \\
  \hline
\end{tabular}
\caption{\footnotesize Default parameter setting. I: inductive, T: transductive, AL: aggregation layer, FC: fully-connected layer, SM: softmax layer.
\label{tab:setting}}}
\end{table}

\section{Additional experiments}

\subsection{Input inspection as a defense}
\label{sec:randomized-smoothing}

We build our defense upon Randomized-Smoothing (\rnds)\mcite{NeilGong:BackdoorGNN:2020}.

\vspace{2pt}
{\bf Randomized smoothing -- } \rnds applies a subsampling function $\gS$ over a given graph $G$ (including both its structural connectivity and node features), generates a set of subsampled graphs $G_1, G_2, \ldots, G_n$, and takes a majority voting of the predictions over such samples as $G$'s final prediction. Intuitively, if $G$ is trigger-embedded, the trigger is less likely to be effective on the subsampled graphs, due to their inherent randomness. In particular, $\gS$ is controlled by a parameter $\beta$ (subsampling ratio), which specifies the randomization magnitude. For instance, if $\beta = 0.8$, $\gS$ randomly removes 20\% of $G$'s nodes, and for the rest nodes, randomly sets 20\% of their features to be 0. Note that while in\mcite{NeilGong:BackdoorGNN:2020}, \rnds is further extended to mitigate trojan GNNs, here we focus on its use as a defense against trigger-embedded graphs.

\begin{figure}[!ht]
    \centering
    \epsfig{file = 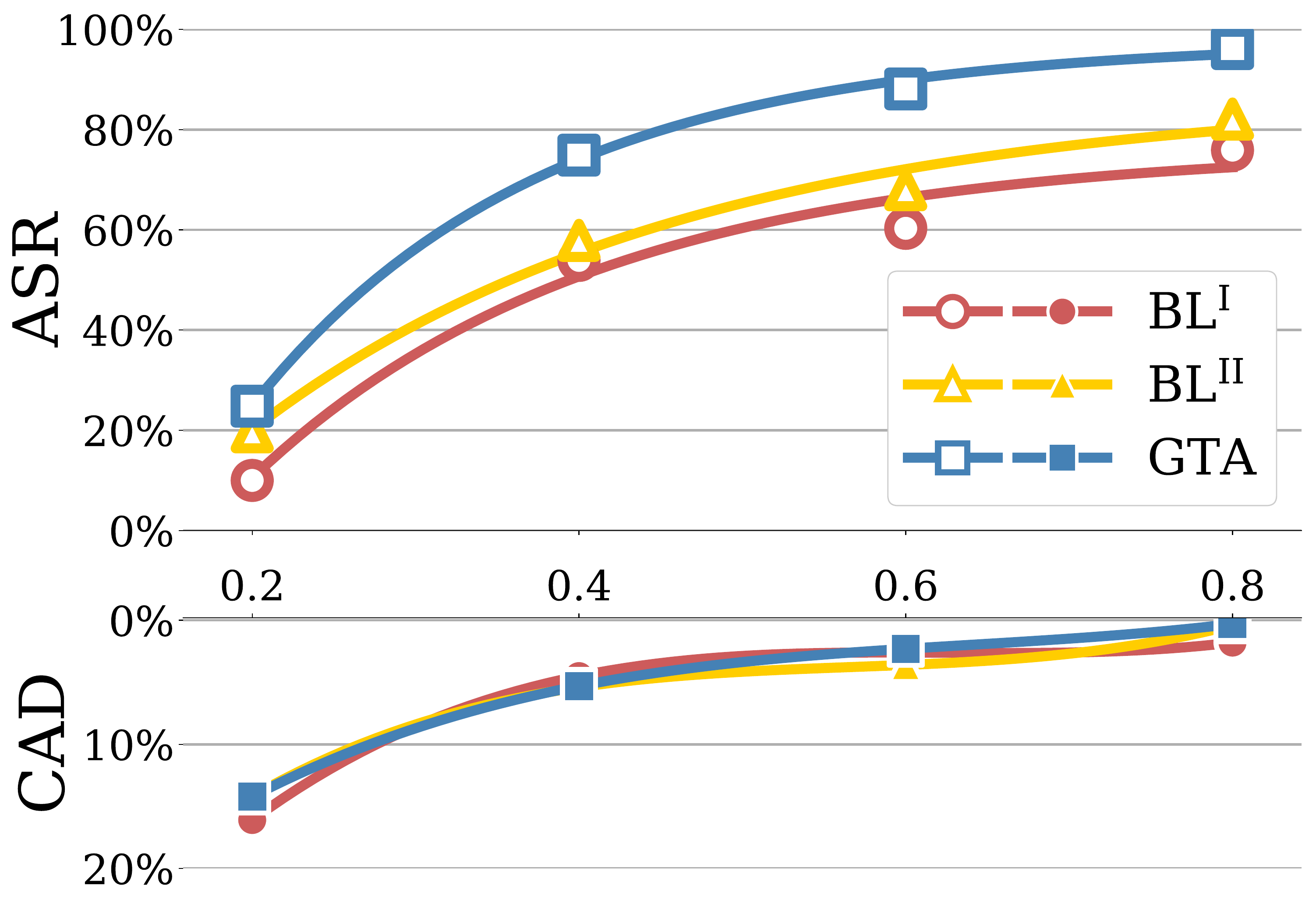, width = 55mm}
    \caption{\footnotesize Attack effectiveness and evasiveness of \system with respect to varying subsampling ratio $\beta$. \label{fig:rand_smooth_tox21}}
    \label{fig:rand_smooth_tox21}
\end{figure}

\vspace{2pt}
{\bf Results and analysis} -- We evaluate \rnds in the transfer case of ChEMBL$\rightarrow$\toxicant ({\em cf.} Table\mref{tab:wild}). Figure\mref{fig:rand_smooth_tox21} illustrates the effectiveness and evasiveness of \system as a function of the subsampling ratio $\beta$. Observe that there exists an intricate trade-off between attack robustness and clean accuracy. A smaller $\beta$ leads to lower \asr but also results in larger \bad. Therefore, \rnds may not be a viable option for defending against \system, as it may negatively impact system performance.

\vspace{2pt}
{\bf Other input-inspection defenses -- } One may suggest using other input inspection methods. Yet, it is often challenging to extend such defenses from continuous domains (\meg, images) to discrete domains (\meg, graphs). For instance, {\sc Strip}\mcite{Doan:2020:arxiv} is a representative input inspection defense. Intuitively, if an input is embedded with a trigger, its mixture with a benign input is still dominated by the trigger and tends to be misclassified to the target class, resulting in relatively low entropy of the prediction. Unfortunately, it is intrinsically difficult to apply {\sc Strip} to graph-structured data. For instance, it is challenging to meaningfully ``mix'' two graphs. 

\subsection{Input-space attacks}
\label{sec:}


\rev{Table\mref{tab:android_feature_desc} summarizes 40 features associated with each node in the Android call graphs.}

\begin{table}[!ht]{\footnotesize
\setlength\extrarowheight{1.5pt}
\setlength{\tabcolsep}{1.5pt}
\centering
\begin{tabular}{c|l|c|l}
  Feature \# & Definition & Value & Constraint \\
  \hline
  \hline
  \multirow{2}{*}{0} & \multirow{2}{*}{number of parameters} & \multirow{2}{*}{$[0, 20]$} & increasing only, requiring\\
  & & & to modify 1$\sim$20 accordingly \\
  \hline
  1$\sim$20 & parameter type (\meg, `int') & $[0, 23]$ & subtype to supertype only\\
  \hline
  21 & return type & $[0, 21]$ & same as above \\
    \hline
  22 & modifier (\meg, `private') & $[0, 4]$ & increasing only \\
  \hline
  23$\sim$37 & instruction call frequency & $[0, \infty]$ & increasing only \\
    \hline
  38 & affiliation type & $[0, \infty]$ & non-modifiable \\
    \hline
  39 & package name & $[0, \infty]$ & non-modifiable \\
  \hline
\end{tabular}
\caption{\footnotesize Descriptive features of Android call graphs and corresponding perturbation constraints.
\label{tab:android_feature_desc}}}
\end{table}

\begin{figure}[!ht]
\centering
  \epsfig{file = 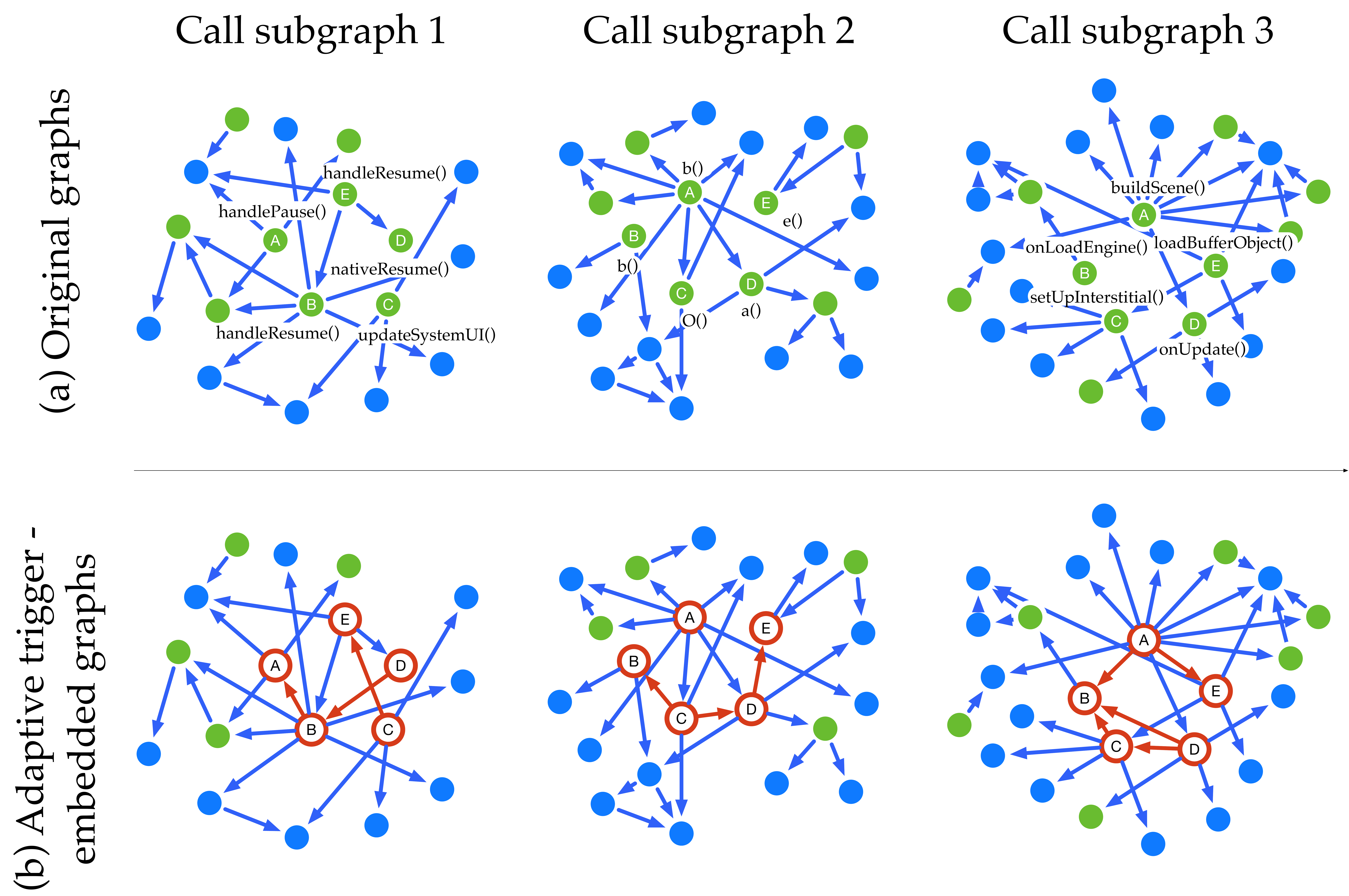, width = 82mm}
  \caption{\footnotesize Illustration of input-space \system on Android call graphs: green node -- external method; blue node -- Android internal method; red node -- method perturbed by \system; blue edge -- original call; red edge -- no-op call added by \system. Only the vicinity of the perturbed subgraph is shown.
  \label{fig:android_callgraph}}
\end{figure}

\rev{Figure\mref{fig:android_callgraph} visualizes sample call graphs generated by input-space \system, where only external methods are perturbed.}

\section{Graph-space constraints}
\label{sec:graph-constraint}

We consider two types of constraints specified respectively on $G$'s topological connectivity and node features respectively.

\vspace{2pt}
{\bf Topological structures --} Let $g$ be the subgraph in $G$ to be replaced by the trigger $\ssub{g}{t}$.
Let $A$ denote $g$'s adjacency matrix with $A_{ij}$ indicating whether nodes $i,j$ are connected. Recall that to generate $\ssub{g}{t}$, \system computes another adjacency matrix $\tilde{A}$ as in \meq{eq:edge-pred} and replaces $A$ with $\tilde{A}$. We consider the following constraints over this operation.

\begin{mitemize}
\setlength\itemsep{0.5pt}

\item Presence/absence of specific edges, which excludes certain pairs of nodes from perturbation. We define the perturbation as:
$\ssub{M}{msk} \odot A + (1 -\ssub{M}{msk}) \odot \tilde{A}$, where $\ssub{M}{msk}$ is a binary mask matrix and $\odot$ denotes element-wise multiplication. Intuitively, $A_{ij}$ is retained if the $ij$-th entry of $\ssub{M}{msk}$ is on and replaced by $\tilde{A}_{ij}$ otherwise.

\item Addition/deletion only which specifies whether only adding/removing edges is allowed. To enforce the addition-only constraint (similar in the case of deletion only), we set the $ij$-th entry of $\ssub{M}{msk}$ to be 1 if $A_{ij}$ is 1 and 0 otherwise, which retains all the edges in $g$.

\item Perturbation magnitude, which limits the number of perturbed edges. To enforce the constraint, we add a regularizer $\| A - \tilde{A} \|_F$ to the objective function in \meq{eq:overall21}, where $\|\cdot\|_F$ denotes the Frobenius norm. Intuitively, the regularizer penalizes a large perturbation from $A$ to $\tilde{A}$.

\item (Sub)graph isomorphism, which dictates the isomorphism of $g$ and $\ssub{g}{t}$ (extensible to other (sub)graphs of $G$ and $\mixfunc$). To enforce this constraint, we add a regularizer $\Delta(\rho(A), \rho(\tilde{A}))$ to the objective function in \meq{eq:overall21}, where $\rho$ maps a graph to its encoding for isomorphism testing and $\Delta$ measures the difference between two encodings. In particular, $\rho$ can be modeled (approximately) as a GNN\mcite{gnn-representation}.

\end{mitemize}

\vspace{2pt}
{\bf Node features --} 
Recall that \system replaces the feature vector $\ssub{X}{i}$ of each node $i \in g$ with its corresponding feature $\ssub{\tilde{X}}{i}$ in $\ssub{g}{t}$. We consider two types of constraints on this operation.
\begin{mitemize}
\setlength\itemsep{0.5pt}
\item Exclusion of specific features, which excludes certain features from perturbation. To improve the trigger evasiveness,
we may restrict the replacement to certain features: $\ssub{v}{\mathrm{msk}} \odot \ssub{X}{i} + (1 - \ssub{v}{\mathrm{msk}})\odot \ssub{\tilde{X}}{i}$, where the mask $\ssub{v}{\mathrm{msk}}$ is a binary vector and $\odot$ denotes element-wise multiplication. Intuitively, the $j$-th feature of $\ssub{\tilde{X}}{i}$ is retained if the $j$-th bit of $\ssub{v}{\mathrm{msk}}$ is on and replaced by the $j$-th feature of $\ssub{\tilde{X}}{i}$ otherwise.

\item Perturbation magnitude, which limits the number of perturbed features. To enforce this constraint, we consider the binary mask $\ssub{v}{\mathrm{msk}}$ as a variable and limit the cardinality of $\ssub{v}{\mathrm{msk}}$ by adding a regularizer $\|\ssub{v}{\mathrm{msk}}\|_1$ to the objective function in \meq{eq:overall21}, where $\|\cdot\|_1$ denotes the $\ell_1$ norm.
\end{mitemize}


\end{document}